# Data Harmonisation for Information Fusion in Digital Healthcare: A State-of-the-Art Systematic Review, Meta-Analysis and Future Research Directions

Yang Nan[1*], Javier Del Ser[4,5], Simon Walsh[1], Carola Schönlieb[6], Michael Roberts[6,7], Ian Selby[8], Kit Howard[9], John Owen[9], Jon Neville[9], Julien Guiot[10,11], Benoit Ernst[10,11], Ana Pastor[12], Angel Alberich-Bayarri[12], Marion I. Menzel[13,14], Sean Walsh[15], Wim Vos[15], Nina Flerin[15], Jean-Paul Charbonnier[16], Eva van Rikxoort[16], Avishek Chatterjee[17], Henry Woodruff[17], Philippe Lambin[17], Leonor Cerdá-Alberich[18], Luis Martí-Bonmatí[18], Francisco Herrera[19,20†], Guang Yang [1,2,3†*]

[1] National Heart and Lung Institute, Imperial College London, London, UK

[2] Cardiovascular Research Centre, Royal Brompton Hospital, London, UK

[3] School of Biomedical Engineering & Imaging Sciences, King's College London, London, UK

[4] Department of Communications Engineering, University of the Basque Country UPV/EHU, 48013 Bilbao, Spain

[5] TECNALIA, Basque Research and Technology Alliance (BRTA), 48160 Derio, Spain

[6] Department of Applied Mathematics and Theoretical Physics, University of Cambridge, Cambridge, UK

[7] Oncology R&D, AstraZeneca, Cambridge, UK

[8] Department of Radiology, University of Cambridge, Cambridge, UK

[9] Clinical Data Interchange Standards Consortium, Austin, Texas, United States

[10] University Hospital of Liège (CHU Liège), Respiratory medicine department, Liège, Belgium

[11] University of Liege, Department of clinical sciences, Pneumology-Allergology, Liège, Belgium

[12] QUIBIM, Valencia, Spain

[13] Technische Hochschule Ingolstadt, Ingolstadt, Germany

[14] GE Healthcare GmbH, Munich, Germany

[15] Radiomics (Oncoradiomics SA), Liège, Belgium

[16] Thirona, Nijmegen, the Netherlands

[17] Department of Precision Medicine, Maastricht University, Maastricht, the Netherlands

[18] Medical Imaging Department, Hospital Universitari i Politècnic La Fe, Valencia, Spain

[19] Department of Computer Sciences and Artificial Intelligence, Andalusian Research Institute in Data Science and Computational Intelligence (DaSCI) University of Granada, Granada, Spain

[20] Faculty of Computing and Information Technology, King Abdulaziz University, Jeddah, 21589, Saudi Arabia

[†] Francisco Herrera and Guang Yang are co-last authors of this work.

[*] Corresponding authors: Yang Nan and Guang Yang.

Yang Nan

E-mail: y.nan20@imperial.ac.uk

Phone: +44 02073528121

Guang Yang

E-mail: g.yang@imperial.ac.uk

Phone: +44 02073528121

## Abstract

Removing the bias and variance of multicentre data has always been a challenge in large scale digital healthcare studies, which requires the ability to integrate clinical features extracted from data acquired by different scanners and protocols to improve stability and robustness. Previous studies have described various computational approaches to fuse single modality multicentre datasets. However, these surveys rarely focused on evaluation metrics and lacked a checklist for computational data harmonisation studies. In this systematic review, we summarise the computational data harmonisation approaches for multi-modality data in the digital healthcare field, including harmonisation strategies and evaluation metrics based on different theories. In addition, a comprehensive checklist that summarises common practices for data harmonisation studies is proposed to guide researchers to report their research findings more effectively. Last but not least, flowcharts presenting possible ways for methodology and metric selection are proposed and the limitations of different methods have been surveyed for future research.

# 1. Introduction

Computational biomedical research aims to advance digital healthcare and biomedical studies by developing computational models that improve the precise diagnosis of disease spectrum, analysis of gene expressions or time series data (e.g., electroencephalograms and electrocardiograms). These models are designed to discover novel risk biomarkers, predict disease progression, design optimal treatments, and identify new drug targets for applications such as cancer, pulmonary disease, and neurological disorders. Whilst a well-performed model should have characteristics of high performance, robustness, explainability, and reproducibility, it faces the issue that the bias of distribution between different datasets dramatically increases the difficulty of developing models from large-scale studies. Although data harmonisation is needed with almost any kind of medical data, automated methods have been extensively used for medical images, gene expression analysis, with the rest of the modalities being ignored or harmonised manually. Studies have shown that machine learning based approaches, especially deep neural networks, are highly sensitive to the distribution of training data. Therefore, there is an urgent need to develop approaches that can integrate the device/site-invariant information from multiple datasets. To address this issue, researchers established standard acquisition protocols [1-3] or definitions [4, 5] to help data collectors to glean standardised data. For instance, Delbeke et al. [2] recommended an acquisition protocol for F-FDG Positron emission tomography/computerised tomography imaging (PET/CT), and Simon et al. [3] presented a standardised MR imaging protocol for multi-sclerosis. Schmidt et al. [4, 5] mainly focused on integrating the data from routine health information systems, including conducting manual harmonisation and rule-based alignment of electronic data. Although these acquisition protocols could effectively reduce the cohort bias (non-biological variances in cross-scanner/site data), they were limited in assisting prospective studies because most studies were retrospective and could not be re-acquired with the same standard. In addition, a non-standardised acquisition protocol is needed for personalised digital healthcare sometimes. Therefore, it is imperative to explore a computational method to harmonise multicentre datasets.

Table 1. Comparison of existing data harmonisation review studies.

| Survey | [6] | [7] | [8] | [9] | **Ours** |
|---|---|---|---|---|---|
| Period | ~2020 | ~2019 | ~ 2020 | ~2021 | ~2021 |
| # of reviewed studies | N/A | 23 | 49 | 42 | 96 |
| Domain | Radiomics | Radiomics | Radiomics | Radiomics | Radiomics, Gene, Pathology |
| Metric | × | × | × | × | √ |
| Checklist | × | × | × | × | √ |
| Guidance | × | √ | × | × | √ |
| Meta-analysis | × | √ | × | × | √ |

"# of reviewed studies" indicates the number of included papers in the survey.

Although some surveys of computational data harmonisation have been released [6, 7, 10], such as MRI (magnetic resonance imaging) [11] or CT (computerised tomography) harmonisation, these surveys only explored methods of single modality or application and rarely focused on evaluation metrics and research guidance (shown in Table 1). Moreover, there is a lack of a checklist that can summarise the common practice and give guidance for methodology selection and development for computational data harmonisation studies. This survey summarises the computational data harmonisation strategies for multimodal data in the digital healthcare field in terms of methodologies, evaluations, and applications. Our paper covers three main areas (i.e., gene expression, radiomics, and pathology), with over 96 qualified papers published within two decades. This is the largest and the most comprehensive exploration of the computational data harmonisation strategies to the best of our knowledge. To provide a better scientific practice for the community working on data harmonisation, a comprehensive checklist with all the steps is proposed to guide the researchers on reporting their studies more effectively. With this checklist, explorations (what the strategy is) and advances (how well the model performs) of the study can be clearly illustrated by reporting the items in model and evaluation sections. Overall, the main contributions of this survey can be summarised as:

- A three-fold taxonomy that describes the methodology, evaluation and applications of computational data harmonisation strategies.
- A checklist with all the steps that can be followed in future data harmonisation studies.
- The critique and limitations of the existing data harmonisation strategies and potential studies.

The rest of the manuscript is organised as follows: (1) section 2 describes the definition, motivation, utilisation and solution of computational data harmonisation issues; (2) section 3 illustrates how this survey is conducted; (3) sections 4, 5, and 6 demonstrate the three-fold taxonomy of harmonisation strategies; (4) section 7 describes the results of the meta-analysis and presents a checklist for data harmonisation studies; (5) section 8 presents the checklist for harmonisation studies and summarises the critiques and limitations of current strategies; and (6) section 9 concludes this survey.

# 2. Computational data harmonisation: definition, origin, what for and how?

This section illustrates the details of data harmonisation, including the definition, origin, purpose and solutions of computational data harmonisation tasks. To better describe these characteristics, the terminology of computational data harmonisation is illustrated in Table 2.

Table 2. Terminology of computational data harmonisation.

| Terminology | Definitions |
|---|---|
| Cohort | A group of data acquired by the same acquisition protocol and devices |
| Subjects | Patients (objects) involved in the study |
| Category | The classes that were involved in the study, e.g., cancer vs. normal |
| Cases | Samples (a subject can produce multiple samples with different acquisition protocols) involved in the study |
| Cohort bias | The non-biological related variances caused by acquisition protocols (also named as "batch effect" in gene expression studies) |
| Source cohorts | The cohort that needs to be harmonised from |
| Reference cohort | The cohort that needs to be harmonised to |

## 2.1 What?

Data harmonisation refers to combining the data from different sources to one cohesive data set by adjusting data formats, terminologies and measuring units [12]. It is mainly performed to address issues caused by nonidentical annotations or records of different operators or systems, which requires a standard protocol for manual adjustment. The conventional approach for data harmonisation is performed by manually setting rules or terms to integrate multicentre datasets from health information systems. It requires complex mapping of terminologies and manual harmonisations.

Different from manual harmonisation that relies on a standard protocol and manual adjustment, computational data harmonisation in digital healthcare aims to reduce the cohort bias (non-biological variances) given by different data acquisition schemes. It applies computational strategies (such as machine learning, image/signal processing) to integrate multicentre datasets and reduce their non-biological heterogeneity. Compared with data cleansing, data normalisation, standardisation, etc., data harmonisation has a broader definition and is a term that represents the strategies of reducing cohort biases (caused by different acquisition protocols and devices). It can be conducted by removing outliers (data cleansing), aligning the location-and-scale parameters of cohorts (data normalisation), converting multiple datasets into a common data format (data standardisation/transformation, referring to manual harmonisation). It is of note that data harmonisation is not as same as style transformation (e.g., generating T1 images using T2 in MRI, or generating CT using X-Ray images), it only focuses on intra-modality datasets.

## 2.2 Why?

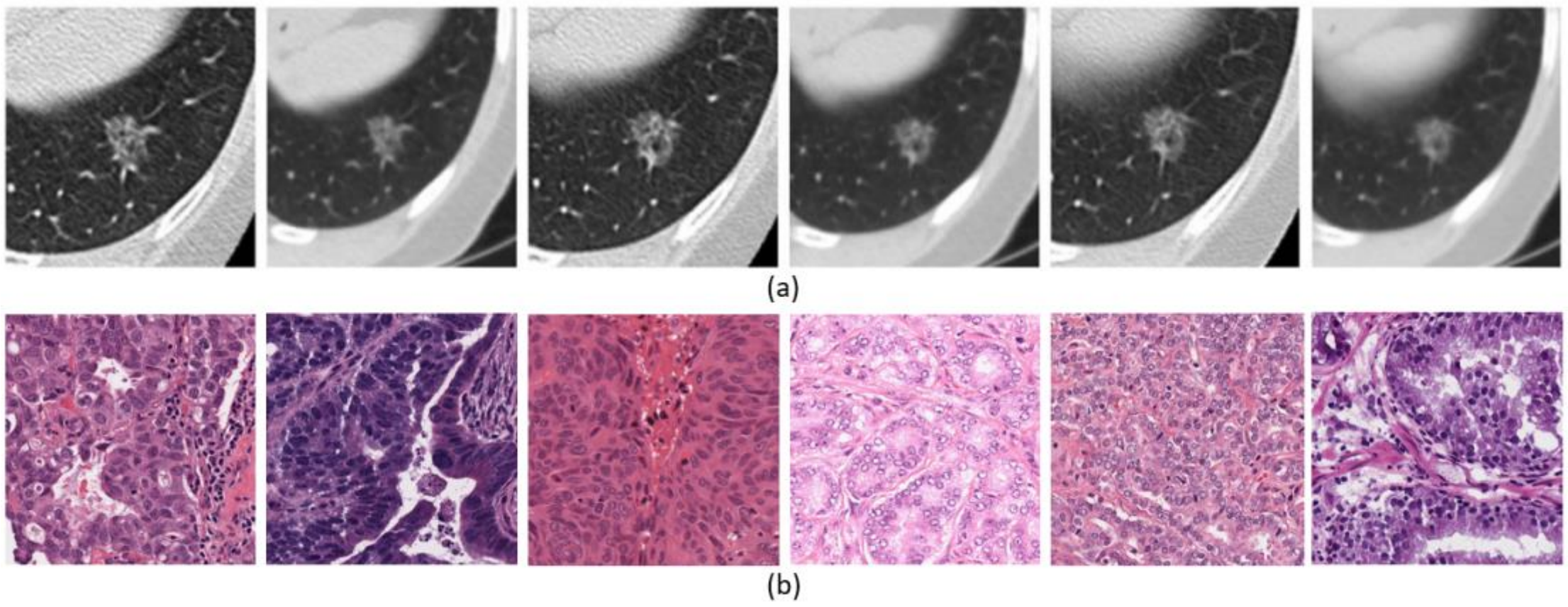

Fig. 1 Visualised differences in (a) radiomics and (b) pathology images. (a) a lung tumour captured on the same CT scanner with 6 different acquisition protocols (From [13]). (b) H&E stained tissue images from different sites [14].

This section first illustrates the motivation of computational data harmonisation approaches, then describes the source of non-biological variances. Computational methods refer to the automatic analysis of digital healthcare data, using machine learning or mathematical modelling algorithms. It usually requires the extraction and fusion of data-derived features from the raw data. For instance, the grey level co-occurrence matrix (GLCM), which is one of the most commonly used textual features in radiomics, can be used as an independent prognostic factor (representing in F-FDG PET/CT images the metabolic intra-tumoral heterogeneity) in patients with surgically treated rectal cancer [15]. However, datasets acquired from different sites present significant variances (Fig. 1), which can hinder the effectiveness of extracted features and lead to unstable performance for both computational and manual diagnosis. In particular, Zhao et al. [13] found a considerable segmentation based inconsistency of lung tumours while conducting repeated manual labelling by three radiologists. This inconsistency could lead to a significant reduction (from 0.76 to 0.28) of concordance correlation coefficients for certain radiomics features. Therefore, computational data harmonisation is proposed to eliminate or reduce these non-biological variances in multicentre datasets for (1) enhancing the robustness and reproducibility of computational modules; (2) producing the fusion of knowledge captured beforehand with knowledge captured over a new task; (3) promoting the comprehensive performance of computational modules.

The non-biological data variances are mainly from hardware (e.g., scanners and platforms), acquisition protocols (e.g., signal/imaging acquisition parameters) and laboratory preparations (e.g., staining and slicing). These variances may lead to the weak reproducibility of quantitative biomarkers and limit the time-series studies based on multi-source datasets, indicating an urgent need for data harmonisation strategies to generate reproducible features [15, 16].

**Heterogeneity of acquisition devices (inter-device variability)**

Heterogeneity of acquisition devices leads to the variance of multicentre data, which is mainly discovered in signals, CT, MRI, and pathological images. This heterogeneity is mainly brought by different detector systems of vendors, the sensitivity of the coils, positional and physiologic variations during acquisition, and magnetic field variations in MRI, among others. [17-20]. Studies have shown that even using a fixed acquisition protocol for different brands of scanners, some radiomics features are still non-reproducible. For instance, Berenguer et al. [21] explored the reproducibility of radiomics features on five different scanners with the same acquisition protocol and witnessed large differences, ranging from 16% to 85% of the radiomics features were reproducible. Sunderland et al. [22] explored the large variance of standard uptake value (SUV) in different brands of scanners, witnessing a much higher maximum SUV of newer scanners compared with old ones.

Table 3. Summary of the reproducibility/repeatability studies.

| Reference | Intra-repro | Inter-repro | Repeatability | Condition | Variables | Object | Modality |
|---|---|---|---|---|---|---|---|
| Jha et al. [23], 2021 | 30.7% (332/1080) | 14.3% (154/1080) | 82.2% (888/1080) | ICC >0.90 | Slice Sickness | Phantoms | CT |
| Emaminejad et al. [24], 2021 | 8.0% (18/226) | / | / | CCC>0.90 | Reconstruction | Patients | CT |
| | 7.5% (17/226) | / | / | CCC>0.90 | Radiation Dose | Patients | CT |
| Kim et al. [25], 2021 | 11.0% (112/1020) | / | / | CCC>0.85 | Acceleration Factors | Patients | MRI |
| Ymashita et al. [26], 2020 | / | 5.6% (15/266) | / | CCC>0.90 | Different Scanners | Patients | CECT |
| Fiset et al. [27], 2019 | / | 22.6% (398/1761) | / | ICC >0.90 | Different Scanners | Patients | MRI |
| Saeedi et al. [28], 2019 | 20.5% (8/39) | / | / | CoV< 5% | Tube Voltage | Phantoms | CT |
| | 30% (13/39) | / | / | CoV< 5% | Tube Current | Phantoms | CT |
| Meyer et al. [29], 2019 | 20.8% (22/106) | / | / | $R^2 > 0.95$ | Radiation Dose | Patients | CT |
| | 52.8% (56/106) | / | / | $R^2 > 0.95$ | Reconstruction | Patients | CT |
| | 39.6% (42/106) | / | / | $R^2 > 0.95$ | Reconstruction | Patients | CT |
| | 12.3% (13/106) | / | / | $R^2 > 0.95$ | Slice Sickness | Patients | CT |
| Perrin et al. [30], 2018 | 24.8% (63/254) | / | / | CCC>0.90 | Injection Rates | Patients | CECT |
| | 13.4% (34/254) | / | / | CCC>0.90 | Resolution | Patients | CECT |
| Midya et al. [31], 2018 | 11.7% (29/248) | / | / | CCC>0.90 | Tube Current | Phantoms | CT |
| | 19.8% (49/248) | / | / | CCC>0.90 | Noise | Phantoms | CT |
| | 63.3% (157/248) | / | / | CCC>0.90 | Reconstruction | Patients | CT |
| Altazi et al. [32], 2017 | 21.5% (17/79) | / | / | Mean difference <25% | Reconstruction | Patients | PET |
| Zhao et al. [13], 2016 | 11.2% (10/89) | / | / | CCC>0.90 | Reconstruction | Patients | CT |
| | / | / | 69.7% (62/89) | CCC>0.90 | / | Patients | CT |
| Hu et al. [33], 2016 | / | / | 64.0% (496/775) | ICC>0.80 | / | Patients | CT |
| Choe et al. [34], 2019 | 15.2% (107/702) | / | / | CCC>0.85 | Reconstruction | Patients | CT |

CCC: concordance correlation coefficient; ICC: intraclass correlation coefficient; CoV: coefficient of variation; $R^2$: R-squared; CT: computed tomography; MRI: magnetic resonance imaging; CECT: consecutive contrast-enhanced computed tomography; PET: positron emission tomography.

### Heterogeneity of acquisition protocols (intra-device variability)

The different acquisition protocols are the main reasons for cross-cohort variability. They mainly include the scanning parameters (e.g., voltage, tube current, the field of view, slice thickness, microns per pixel, etc.) and reconstruction approaches (e.g., different reconstruction kernels) [35]. To investigate the intra/inter reproducibility of radiomics features, several studies have been conducted by test-reset experiments (Table 3). In Table 3, a good reproducibility/repeatability is defined as the high correlation coefficient (e.g., ICC, CCC, $R^2$) or low difference (e.g., mean difference, CoV) between two features. For instance, a certain radiomics feature is considered reproducible/repeatable when the CCC between features extracted from two repeated scans is larger than 0.90. As shown in Table 3, the scanning parameters notably affect the radiomics features, making the statistical analysis difficult. For instance, only 15.2% of radiomics features are reproducible when using soft and sharp kernels during the reconstruction [34]. This weak reproducibility greatly hinders the large-scale digital healthcare studies and applications of computational models. Although implementing strict standard protocol can reduce non-biomedical variances, the non-standard acquisition protocol is needed by physicians for personalised centre-based image quality considerations. For instance, the thickness and pixel size are regularly adjusted on a case-by-case principle to improve the data quality [36]. Therefore, the heterogeneity of acquisition protocol is unavoidable which requires a general solution.

**Heterogeneity of laboratory preparations (Preparation variability)**
All the gene expression, radiomics, and pathological data heavily suffer from laboratory variances, including sample preparation, assay, slicing, and staining. For single-cell RNA sequencing (scRNA-seq) and microarray data, there are various analysis platforms with different biases, making it difficult to integrate and compare results from multi-centre/batch of data [37, 38]. For radiomics data, variances such as injection rate and radiation dose may also affect the data quality. Considering the pathology data, variances are mainly from manual operations [39, 40] (e.g., biopsy sectioning, sample fixation, dehydration and stain concentration), all these factors result in the variation of pixel values and stain consistencies.

## 2.3 What for?

**Large scale and longitudinal studies.** The challenges of integrating and utilising multicentre datasets make researchers realise the importance of data harmonisation when conducting large-scale studies [41]. On the one hand, the information fusion without harmonisation cannot achieve reproducible results in large scale and longitudinal studies [13, 31, 42]. Some researchers have advised that the conclusions reached must be treated with caution since some features can vary greatly against minor non-biomedical changes [43]. Data harmonisation, on the other hand, is critical for patients who are monitored longitudinally and imaged on different scanners. For instance, the longitudinal PET cannot provide helpful information if they are gathered from multi-scanners, since the relationship between SUV and outcomes may get concealed [16].

**Transferability of computational models.** The unstable performance has been found when applying computational models to multicentre datasets [44]. To address this issue, transfer learning was proposed to enhance the robustness of computational models by holding a priori knowledge on the way data can vary. It feeds the model with further data which reflects the variability that the model may encounter at inference time. However, transfer learning requires extra training samples to reduce the uncertainty with respect to the variability of data that models can cope with. This could be inapplicable for prospective studies in the digital healthcare field. Different from transfer learning, computational data harmonisation strategies can process the data without extra training or fine-tuning, which provide an applicable solution for multicentre studies. Meanwhile, there has been mounting evidence that combining data harmonisation with machine learning algorithms enables robust and accurate predictions on multicentre datasets [45].

## 2.4 How?

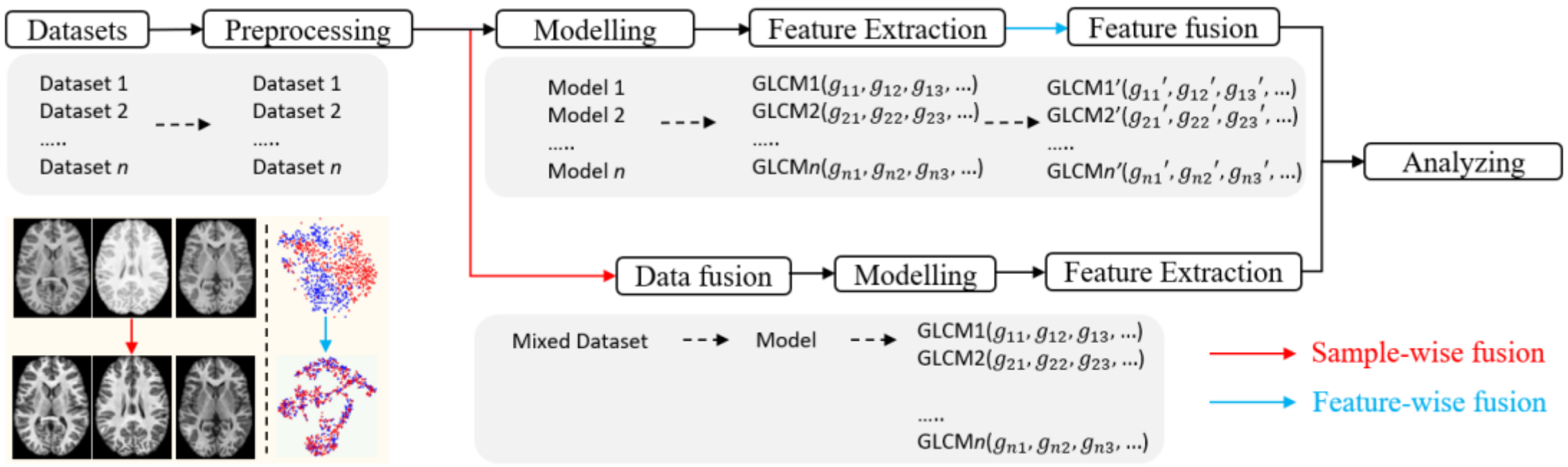

Fig. 2 Workflow of developing a computational data harmonisation method.

The deployment of a computational method includes preparation (acquiring datasets such as staining, scanning), pre-processing, modelling and analysing, while the data harmonisation can be performed through the processing of images/signals/gene matrices (i.e., sample-wise) or alignment of data-derived features (i.e., feature-wise). The sample-wise harmonisation is usually conducted before modelling, aiming to reduce the cohort variance of all training samples and fuse multicentre samples as a single dataset. It involves image processing, synthesis and invariant feature learning approaches. After acquiring cohort-invariant data, a single model can be developed for clinical related tasks. The feature-wise harmonisation aims to reduce the bias of extracted features, such as the GLCM, convex hull area of the region of interest. It is usually performed on extracted feature matrixes, eliminating the cohort variances through fusing the extracted features (shown as the left bottom subfigure Fig. 2, the red and blue dots indicate samples from different cohorts). Both the sample-wise and feature-wise data harmonisation can effectively reduce the variances and improve the performance of the analysis. However, the feature-wise harmonisation requires several models to extract features of interest, leading to complex model development. Moreover, when the number of samples in each cohort is small, it is hard to develop the corresponding models.

# 3. Methods

## 3.1 Literature search and review

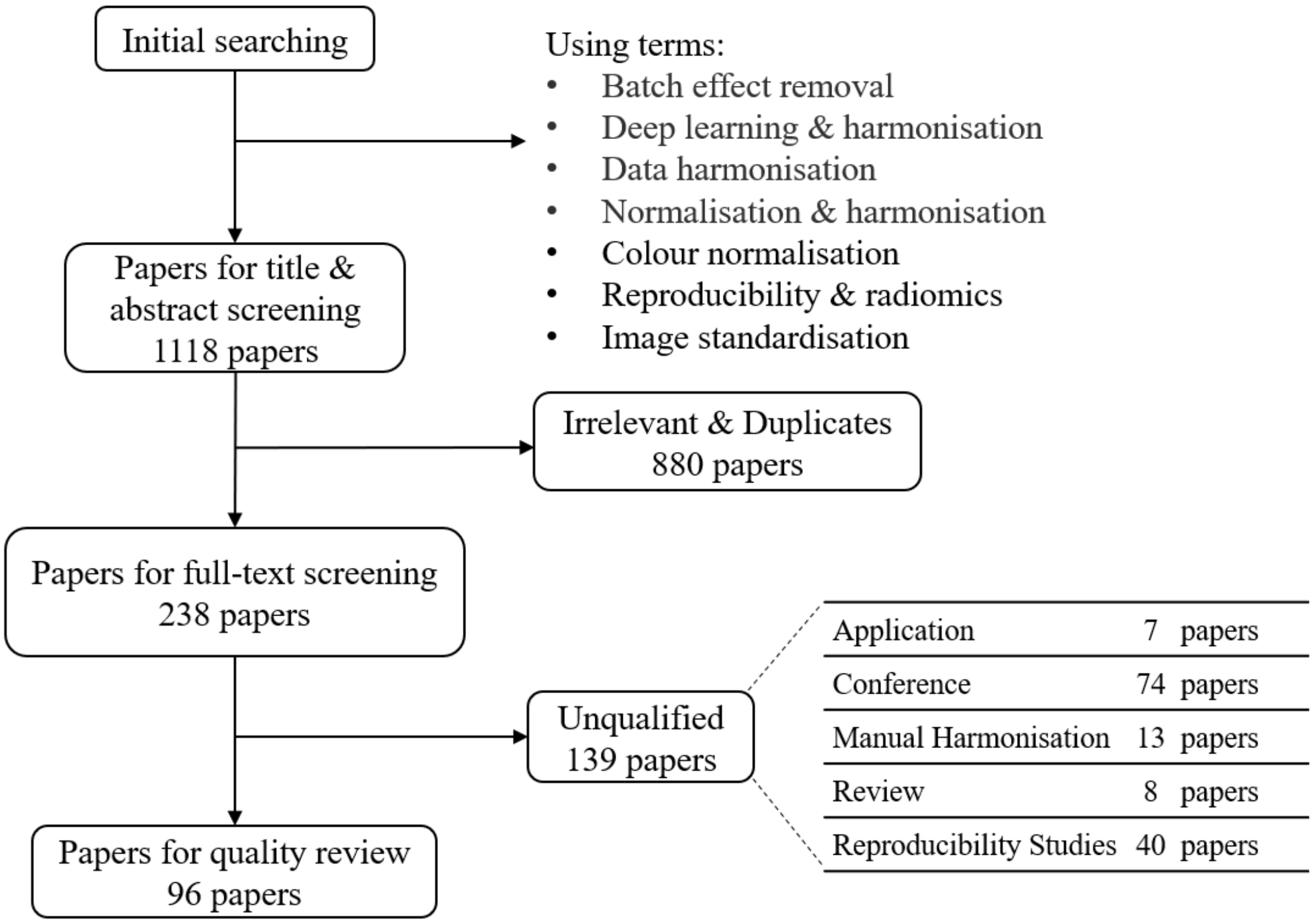

Fig. 3 Literature selection procedure.

The literature search, selection and recording were conducted independently by two researchers with experience in computer science and biomedicine. The agreement was then achieved by a third reviewer with the expertise of biomedical data analysis. All these searches were performed on Scopus Preview (Elsevier) database for publications up to July 10, 2021. To investigate the strategies of harmonisation for information fusion, we searched the literature using the keyword of 'batch effect removal', 'deep learning' and 'harmonisation', 'data harmonisation', 'normalisation' and 'harmonisation', 'colour normalisation', 'reproducibility' and 'radiomics', 'image standardisation'. These initial keywords were searched both independently and jointly to cover more literature. It is of note that both 'normalisation' and 'standardisation' are methods of harmonisation. The pre-screening was first conducted by viewing the abstract and title to filter those irrelevant articles. The eligibility was then checked through our criteria (given in section 3.2) to remove the unqualified works for full-text review.

A flowchart demonstrating the literature selection procedure is presented in Fig. 3. After removing the irrelevant and duplicated articles by screening the titles and abstracts, 238 articles were selected for full-text screening. Based on eligibility criteria, 139 publications were considered unqualified, and 96 papers were included in this systematic review.

## 3.2 Inclusion and exclusion criteria

The entry criteria were: (1) original research publications in peer-reviewed journals or international conferences; (2) focus on the computational data harmonisation of digital healthcare data. The excluded criteria were: (1) studies that only applied existing harmonisation strategies without further development; (2) studies that focused on manual harmonisation such as regulations; (3) review and literature survey studies; (4) studies that only explore the reproducibility or stability without developing harmonisation approaches.

### 3.3 Data collection

Details of papers for quality review were manually summarised in a spreadsheet, including title, modality, methodology, metrics, data scale, year of publication, data property (e.g., private or public), applications, number of cohorts, and number of cases.

# 4. Data harmonisation strategies for information fusion

In this systematic review, data harmonisation approaches were divided into four groups, with the distribution based methods, image processing, synthesis, and invariant feature learning. To better illustrate the basic idea and relationship of computational approaches, a taxonomy is shown in Fig 4, followed by a detailed description of harmonisation techniques.

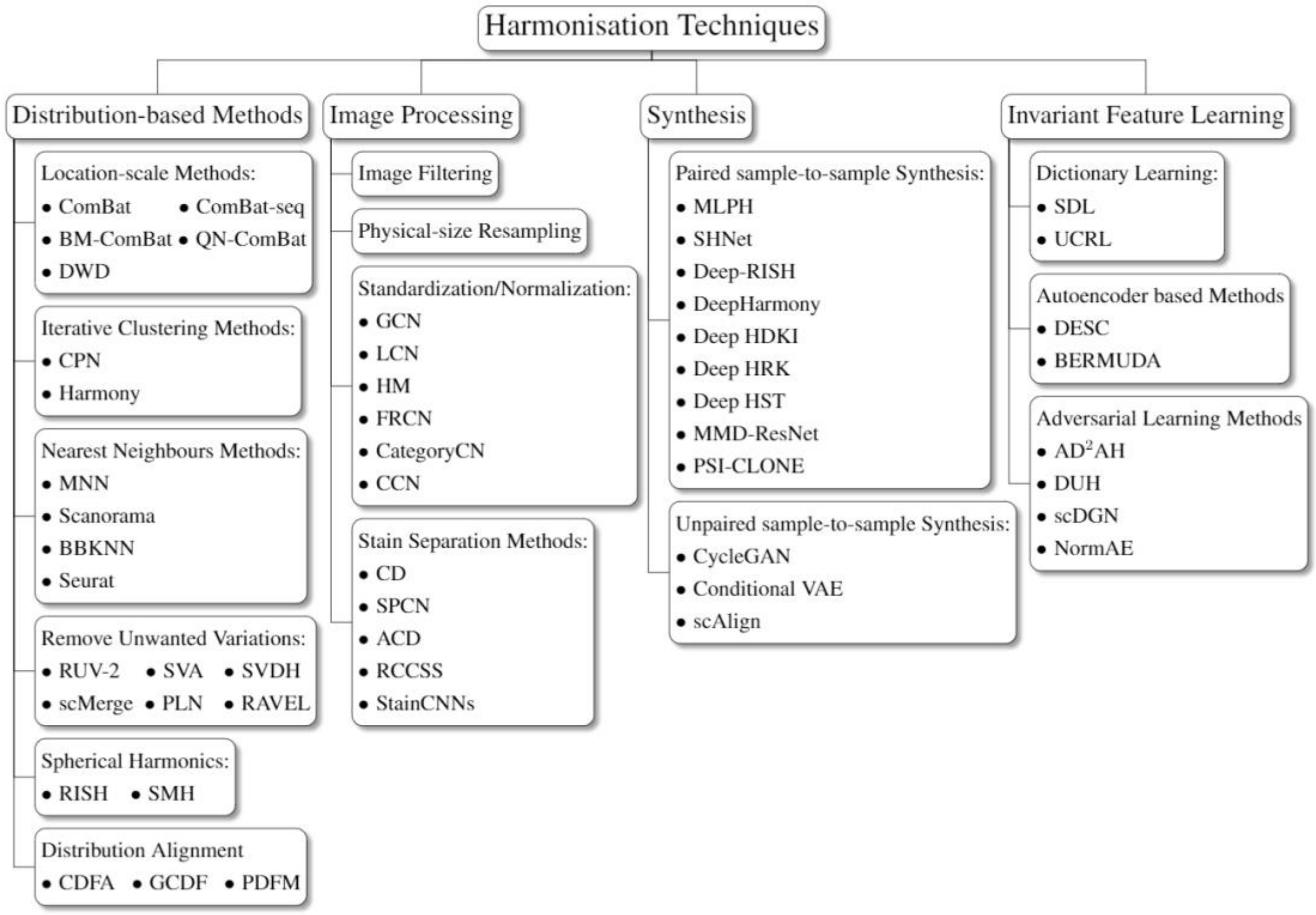

Fig. 4 Taxonomy of computational data harmonisation strategies.

## 4.1 Distribution based methods

The distribution based methods estimate/calculate the bias between cohorts from the latent space, then match/map the source data to the target ones through a bias correction vector or alignment functions.

### 4.1.1 Location-scale methods (LS)

The location-scale methods estimate the location-scale parameters (mean and variance) of each cohort and align all data towards the same location-scale.

**ComBat**: ComBat [46] robustly estimated both the mean and the variance of each batch using empirical Bayes shrinkage, then harmonised the data according to these estimates. The data was first standardised to have similar overall mean and variance, followed by the empirical Bayes estimation via parametric empirical priors. With these adjusted bias estimators, the data could be harmonised by the location-scale model based functions [47-66]. For instance, Radua et al. applied ComBat to address the heterogeneity of cortical thickness, surface area and subcortical volumes caused by various scanners and sequences [53]. Whitney et al. implemented ComBat to harmonise the radiomic features extracted across multicentre DCE-MRI datasets [54].

**ComBat-seq**: Researchers have made more extensions based on the original ComBat harmonisation.

Since the assumption of Gaussian distribution in the original ComBat made it sensitive to outliers, Zhang et al. proposed ComBat-seq [67] by assuming the Negative Binomial distribution, which could better address the outlier issues. The ComBat-seq first built a negative binomial regression model and obtained the estimators of cohort bias, followed by the calculation of 'batch free' distributions for mapping original data.

**BM-ComBat:** Different from the original ComBat that shifted samples to the overall mean and variance, an M-ComBat [68] was proposed to provide a flexible solution, transferring the data to the location and scale of a pre-defined "reference". With these efforts, Da-ano et al. [69] proposed a BM-ComBat by introducing a parametric bootstrap in M-ComBat for robust estimation, aiming to provide a more flexible and robust harmonisation strategy.

**QN-ComBat:** Müller et al. [70] applied a quantile normalisation before ComBat correction in longitudinal gene expression data to achieve better performance.

**Distance-Weighted Discrimination (DWD):** DWD [71] searched the hyperplane where the samples could be well separated and projected the different batches on the DWD plane. The data was then harmonised by subtracting the DWD plane multiplied by the batch mean. It is of note that DWD repeated the translations of samples from different cohorts until their vectors were overlapped.

### 4.1.2 Iterative clustering methods (IC)

The iterative clustering methods harmonise the cohort bias by conducting multiple bias correction through repeated clusterings procedures. These methods usually (1) perform cluster to all samples from different cohorts, and (2) compute the correction vectors for harmonisation based on cluster centroids.

**Cross-platform normalisation (XPN):** XPN [72] took the combined standardised sample and median central gene as input to remove gross systematic differences, followed by the clusters, aiming to identify homogenous groups of genes and samples with similar expressions in combined data. The gene clusters were then acquired by assignment function, which was used to compute estimated model parameters via standard maximum likelihood.

**Harmony:** Harmony [73] first employed principal components analysis (PCA) to reduce the dimension of all samples, and classified them into several groups (one centroid per group) through k-means clustering. With these centroids, the correction factors for harmonisation were calculated. The above clustering and correction were repeated until the convergence.

### 4.1.3 Nearest neighbours methods (NNM)

NNM methods first found the mutual nearest pairs, then computed the bias correction vectors based on paired samples and subtracted these vectors from the source cohort. Differences in these methods mainly refer to the geometry space when locating the mutual nearest pairs.

**Mutual nearest neighbours (MNN):** MNN identified nearest neighbours between different cohorts and treated them as anchors to calculate the cohort bias [74]. It first pre-normalised the gene data with cosine normalisation, followed by the estimation of the bias correction vector by computing the Euclidean distances between paired samples. The bias correction vector was then applied to all samples instead of the participated pairs. It required that all participated batches must share at least one common type with another.

**Scanorama:** Similar to the MNN method, panorama stitching (Scanorama) [75] aims at estimating cohort bias from samples across batches. It first reduced dimensions of raw data (or source data) using singular value decomposition (SVD). Then an approximate nearest neighbour was adopted to find the mutually linked samples across cohorts. Different from MNN, Scanorama checked the priority of dataset merging within all batches and acquired the merged panorama based on the weighted average of batch correction vectors. At last, the harmonisation was performed with Scanpy [76] workflow.

**Batch balanced k-nearest neighbours (BBKNN):** Initially, BBKNN [77] found the nearest neighbours in a principal component space based on Euclidean distances. Then it built a graph that linked all the samples across cohorts based on the neighbour information. These neighbour sets were then harmonised by uniform manifold approximation (UMAP) [78] algorithms.

**Standard CCA and multi-CCA (Seurat):** Different from other NNM-methods, Seurat [79] performed canonical correlation analysis to acquire the canonical correlation vectors that could project multi-datasets into the most correlated subspace. In this subspace, the mutual nearest pairs were located to compute the bias correction vectors to guide the data integration. When processing multi-cohort datasets (number of cohorts larger than two), the first batch would be set as the reference batch for the correction of the second batch. Then the harmonised second batch would be appended to the reference batch. This repeated procedure stopped when all the batches are harmonised [38, 79].

4.1.4 Remove unwanted variations (RUV)

These methods assumed that the cohort bias was independent of those biases refer to biological variances, which could be estimated as “unwanted variations”. For instance, the bias of negative control genes (prior known genes that would not be affected by biological changes of interest) could be regarded as cohort bias. Based on this assumption, the raw data could be harmonised by subtracting those “unwanted variations”.

**Remove unwanted variations, 2-step (RUV-2):** Control variables were used by RUV-2 to discover the factors related to cohort bias [80]. The negative control (probes that should never be expressed in any sample) samples were subjected to component analysis, and the resulting factors were incorporated into a linear regression model. Variations in the expression levels of these genes thus were considered undesirable. To extract low-dimensional features, Risso et al. [81] presented an extension of the RUV-2 with a zero-inflated negative binomial ~~(ZINB)~~ model that accounted for dropouts, discretisation, and the count character of the data. The cohort bias was then subtracted from the raw data to generate a gene expression matrix that is harmonised.

**Singular value decomposition harmonic (SVDH):** By factorising the expression matrix of input data and reconstructing it while taking off the elements related to the cohort bias, singular value decomposition (SVD) could be used to reduce cohort bias. Alter et al. [82] suggested using SVD to harmonise the data by filtering away the eigenarrays that lead to noise or experimental artefacts.

**scMerge:** scMerge [83] first constructed a graph that connected clusterings between cohorts by searching for mutual nearest neighbours. The unwanted factors were then estimated using stably expressed genes as negative controls. At last, an RUV model was used to collect and remove unwanted differences between cohorts.

**Surrogate variable analysis (SVA):** SVA [84] aimed to recognise and estimate the unwanted variations of data from multiple cohorts. It could be performed without any cohort information. The mixed dataset was first divided into a collection of *n* surrogate variables via SVD, followed by the clearance of data with large variances. SVA coefficients were then calculated for harmonisation by using a linear regression function with surrogate variables and raw diffusion intensities.

**Print-tip loess normalisation (PLN):** PLN [85] was initially proposed to deal with microarray data. To eliminate the cohort bias, PLN employed a blocking term to construct a linear model with the input data. The cohort bias was subtracted from the original data to produce the batch corrected expression matrix.

**Removal of artificial voxel effect by linear regression (RAVEL):** RAVEL [86] separated the voxel value into unwanted variation parts and biological parts. The unwanted variation factors were estimated from the region of interest by SVD, based on the prior knowledge of voxel values, which were not related to disease status.

4.1.6 Spherical harmonics (SH)

Spherical harmonics approaches were designed to harmonise MRI data, aiming to coordinate all data from different cohorts to the same spherical harmonic domain, by adjusting the spherical variables.

**Rotation invariant spherical harmonics (RISH):** RISH was based on mapping diffusion-weighted imaging data from source cohorts to target cohorts [17, 66, 87, 88]. It started with calculating the rotation-invariant features from the estimated spherical harmonics coefficients (of target and source samples, respectively). These rotation invariant features were then mapped from the source cohorts to target cohorts through region-specific linear mapping, followed by the updating of spherical harmonics coefficients. The harmonised diffusion signal was calculated for each subject in source cohorts using the latest spherical harmonics coefficients in target cohorts of gradient directions.

**Spherical moment harmonics.** Due to the insufficient adjustment by location-scale parameters in some cases, researchers proposed the spherical moment method (SMM), which utilised the spherical moments to map the diffusion-weighted images from source cohorts to reference cohorts [89, 90]. SMM matches the spherical mean ($M_1$) and spherical variance ($C_2$) per b-value (the diffusion weighting) by $M_1[T_b] = M_1[f(S_b)]$ and $C_2[T_b] = C_2[f(S_b)]$, where $T_b, S_b$ are data from the target and source cohorts under $b$ shell, respectively. The mapping parameters for harmonising data from different cohorts were acquired by the linear transform $f$.

4.1.7 Distribution alignment (DA)

Distribution alignment methods aim to transform the distribution of the source cohort to that of the reference cohort, using cumulative distribution functions or probability density functions.

**Cumulative distribution functions alignment (CDFA):** CDFA [91] was first proposed for multisite MRI data harmonisation, which aligned the source voxel intensities through an estimated non-linear intensity transformation to match the target cumulative distribution functions. The estimated intensity transformation defined a one-to-one mapping between the voxels in source and target cohorts.

**Gamma cumulative distribution functions alignment (GCDF):** The voxel intensities were re-parameterised using a mixture model of two Gamma distributions that fitted a reference histogram [92]. This reparameterisation was based on the CDF of the Gamma component, which modelled the particular uptake, and constrained the new feature space to [0, 1].

**Probability density function matching:** GENESHIFT [93] estimated the empirical density and measured the distance between probability density functions. GENESHIFT first picked the common genes from different cohorts, then estimated their probability density functions to find the best matching offsets. The harmonised data would be acquired by subtracting the estimated offsets from the source cohorts.

## 4.2 Image processing

Image Processing employs digital image processing algorithms to harmonise multi-cohort data, including image filtering (also called image convolution), registration, resampling and normalisation.

### 4.2.1 Image filtering (IF)

Image filtering (also called convolution) is the process that multiplies two arrays to produce a new array of the same dimension. The 2D second-order Butterworth low-pass filter was found to be able to eliminate cohort bias between CT images with different voxel sizes [94], while the local binary pattern filtering could produce stable and reproducible radiomic features [95].

### 4.2.2 Physical-size resampling (Resample)

Studies have shown that physical size such as pixel/voxel size, mpp (microns per pixel of level 0 in digital pathology) can greatly affect the radiomic/pathological features. This bias can be reduced using bilinear resampling to equalise all the physical sizes [94].

### 4.2.3 Standardisation/normalisation (SN)

Standardisation/normalisation models were designed to reduce the variation and inter-variability in different cohorts by linear transform. These methods usually performed location-scale shifts in image spaces (e.g., HSV, RGB, $\alpha\beta$, illumination spaces, etc.) or image histograms.

**Global colour normalisation (GCN)** transfers the colour statistics from the source to the target images by globally altering the image histogram [96, 97]. A typical representative of GCN is Z-score normalisation, assumed the variable from cohort $i$, subject $j$ as $X_{ij}$, z-score normalisation is conducted through

$$X_{ij} = \frac{X_{ij} - \mu_i}{\sigma_i} \tag{1}$$

where $\mu_i$ and $\sigma_i$ are the mean and standard deviation of each cohort. However, this global alignment may lose some information.

**Local colour normalisation (LCN)** transfers the colour statistics of the specific regions, e.g., ignoring the background regions, from source to target images. In [98], the authors first converted the source and target images from the RGB into the $l\alpha\beta$ space, and then conducted a transformation to harmonise the source image and re-converted it into the RGB space. It is of note that the luminance of background regions is not involved during the processing. This helped the transformation to preserve intensity information within the region of interest while requiring the pre-definition of certain regions.

**Histogram matching (HM):** HM is a method of contrast adjustment using the histogram of images [99]. It adjusts the distribution of images by scaling the pixel values to fit the range of specified histogram (i.e., the target one):

$$f(x, y) = \frac{I_{Tmax} - I_{Tmin}}{I_{Smax} - I_{Smin}}(I_S - I_{Smin}) + I_{Tmin} \tag{2}$$

where $I_T$ indicates the target image and $I_S$ is the source image. Generally, $I_{Tmax}$ and $I_{Tmin}$ are 0 and 255, respectively. For instance, Shah et al. [100] investigated the histogram normalisation on MRI images to harmonise cross-cohort data for multiple sclerosis lesion identification.

**Fuzzy based Reinhard colour normalisation (FRCN):** To decrease the colour variation, Roy et al. [101] applied fuzzy logic to regulate the contrast enhancement in $l$ space to adjust the colour coefficients within the $\alpha\beta$ space.

**Category based colour normalisation (CategoryCN):** To reduce the variance of global colour normalisation, researchers proposed a category based approach for accurate colour normalisation [102]. CategoryCN first classified each pixel by unsupervised approaches from the source and target images, then conducted colour normalisation based on the different classes.

**Complete colour normalisation (CCN):** The complete colour normalisation included the normalisation of illumination and spectrum, one to harmonise the illuminant during imaging and another to reduce spectral variation [39, 103]. CCN estimated the illuminant and spectral matrices from the target cohort, then matched the source illuminant and spectral estimations to the target ones.

4.2.4 Stain separation methods (SS)

Stain separation approaches separated the input images into distinct channels (e.g., the haematoxylin channel, eosin channel, and the background channel for H&E-stained images) to evaluate the stain feature matrix and match these features through certain operations from source to target cohort data. The core concept of stain separation was based on Lambert Beer's law [104] (in the RGB space, stain concentrations are nonlinearly dependent), shown as

$$I_C = I_0 \mathrm{e}^{-\mathrm{OD}_c} \tag{3}$$

where $I_0$ was the value of incident light, and $\mathrm{OD}_c$ was the value of images in optical density (OD) space. Most stain separation methods aimed to factorise the OD values into two matrices as

$$\mathrm{OD}_c = \log\left(\frac{I_0}{I_C}\right) = S * D \tag{4}$$

where *S* was the stain depth matrix and *D* was the stain colour appearance (SCA) matrix.

**Colour deconvolution (CD):** These approaches estimated the concentration of stains in pixel values and normalised the spectral variation in separated stains [105-108]. For example, estimation of the stain matrix was first given by evaluating the proportion of RGB channels within different cohorts, followed by colour deconvolution [106, 107]. The inverse of the staining appearance matrix was multiplied with the optical density space intensity value to get normalised stain channels using non-linear spline mapping.

**Structured-preserving colour normalisation (SPCN):** SPCN assumed that most tissue regions were characterised by the most effective stain among the used stains [109]. It first converted a given RGB image to optical density using the Beer-Lambert Law. After that, SPCN decomposed images into several stain density maps using sparse and non-negative matrix factorization (SNMF), followed by the combination of the stain density map and colour normalisation.

**StainCNNs:** Inspired by SPCN, Lei et al. proposed a deep neural network for stain separation to reduce the computational consumption of SNMF [110]. The proposed stainCNNs approach took the source images as input and learned to generate the stain colour appearance matrix. It significantly reduced the processing time while retaining the high quality of the harmonised images.

**Adaptive colour deconvolution (ACD):** ACD first transferred the input RGB images to optical density space, then performed stain separation with adaptive colour deconvolution matrix to obtain the haematoxylin (H) channel, eosin (E) channel and residual channel [111]. At last, the harmonised images were obtained through recombining the H and E components with a stain colour appearance matrix of target cohorts.

**Rough-fuzzy circular clustering based stain separation (RCCSS):** In RCCSS, stain separation was carried out using an image model based on transmission light microscopy [112]. Initially, each image was transferred to OD space and then decomposed to obtain the SCA matrix and associated stain depth matrix. Maji et al. [113] presented a circular clustering algorithm to find the 'centroid', 'a crisp lower approximation', and the 'fuzzy boundary', which could be integrated by saturation-weighted hue histogram in the HIS colour space.

## 4.3 Synthesis

The objective of synthesis is to precisely reproduce a sample that belongs to a missing modality or domain, which harmonises the multi-cohort datasets. It relaxes harmonisation tasks as style transfer and considers each cohort as a 'style' and transfers all samples to the same 'style'. Based on the characteristics of the training sample, synthesis methods are divided into paired synthesis and unpaired synthesis.

### 4.3.1 Paired sample-to-sample synthesis (P-s2s)

P-s2s methods are trained using paired samples generated from the same object acquired using different protocols. These methods aim to learn the data transfer between source and reference cohorts, which require the repeated acquisition of the same subject under different protocols. Therefore, they can only be applied to radiomic data since the repeated acquisition for the same subject is impossible for gene expression and pathology.

**Multi-layer perceptron harmonic (MLPH):** In 2009, a pilot architecture of the autoencoder-related method was proposed by Cheng et al. [114] to generate the harmonised data by learning the nonlinear transform function.

**Spherical harmonic network (SHNet):** Golkov et al. [115] presented a cascaded fully connected network that employs ReLU and Batch normalisation to harmonise the diffusion MRI scans. Inspired by SHNet, Koppers et al. [116] applied the residual structure to improve the robustness while avoiding overfitting.

**Deep rotation invariant spherical harmonics (Deep-RISH):** Karayumak et al. [117] proposed a deep learning based non-linear mapping approach that utilises RISH features to map the raw signal (dMRI data) between scanners with the same fibre orientations. Deep-RISH was composed of five convolution layers, which took the 9×9 RISH feature patches as the input.

**DeepHarmony:** DeepHarmony was proposed to produce data with consistent contrast within different cohorts [118]. It employed a U-Net based architecture, taking data from the source cohort and producing harmonised data of the target cohort.

**Deep harmonics for diffusion kurtosis imaging (Deep HDKI)**: Tong et al. [119] carried out a concise architecture with three 3D-convolution layers for diffusion kurtosis images (DKI). The paired data was generated using an iterative technique called linear least square and were non-linearly registered to diffusion-weighted images acquired on the target scanner using the computational tools. Then the neural network was trained on the paired samples for harmonisation.

**Deep harmonics for slice thickness (Deep HST):** Park et al. [120] studied the reproducibility of radiomic features in lung cancer under different slice thicknesses and proposed an end-to-end deep neural network to generate harmonised CT data between 1-, 3-, and 5-mm slice thickness.

**Deep harmonics for reconstruction kernel (Deep HRK):** Choe et al. [34] explored the influence of different reconstruction kernels on radiomic features and presented a CNN with residual learning to transfer the data from the soft kernel (B30f) to the sharp kernel (B50f).

**Distribution-matching residual network (MMD-ResNet):** Shaham et al. [121] presented a comprehensive multi-layer perceptron for harmonisation with residual connection [122] and batch normalisation [123] techniques. Given two cohorts of data $X[x_1, x_2, \ldots, x_m] \in \mathrm{D}_1$ and $Y[y_1, y_2, \ldots, y_n] \in \mathrm{D}_2$. The MMD-ResNet aimed to learn a map $\hat{\varphi}: R_d \rightarrow R_d$ by minimising the maximum mean discrepancy [124] between $\hat{\varphi}(X)$ and *Y*. It is of note that this was a 'one-way street' distribution matching for harmonisation and required re-training for inverse transformation.

**Pulse sequence information based contrast learning on neighbourhood ensembles (PSI-CLONE):** PSI-CLONE [125] first calculated sequence parameters $\emptyset_s$ from source cohorts, then applied $\emptyset_s$ to the reference cohorts to produce the source-style data. By training a regression model to learn the nonlinear mapping between synthesised source-style data and reference data, the source cohorts could be harmonised effectively. Based on PSI-CLONE, Jog et al. [126] applied the multi-scale feature extraction to improve the performance.

4.3.2 Unpaired sample-to-sample synthesis (Up-s2s)

Up-s2s approaches generate the harmonised data by cycle-consistent generative adversarial networks or conditional variational autoencoder-decoder, which require sufficient samples and cohort labels from different cohorts for network training.

**Cycle-consistent generative adversarial networks (CycleGAN):** Most synthesis methods of unpaired sample-to-sample translation were based on CycleGAN [127, 128] and its derivatives [62, 129, 130]. In [130], a CycleGAN with Markovian discriminator was applied to harmonise the diffusion tensor data, which was designed to further improve the ability to capture local information.

**Conditional variational autoencoder-decoder (Conditional VAE):** Variational Autoencoder (VAE) is commonly used in data synthesis, dimensional reduction, and feature refinement tasks. It employs an encoding network $E_\theta(z|x)$ to decompose the input high dimensional data *x* into hidden representation *z*, and a decoding network $D_\delta(x|z)$ to reconstruct the raw data *x*, where $\theta$ and $\delta$ are parameters of $E$ and $D$. The conditional VAE modifies the decoder to a conditional decoder $D_\delta(x|z, c)$ that takes the latent variable $z$ and specified cohort $c$ back to a harmonised data $\hat{x}$. By integrating Conditional VAE with the adversarial module, cohort transfer can be performed without paired training samples. Several studies have been proposed using Conditional VAE for data harmonisation, including:

(1) **SH-VAE** [131] performed cohort bias correction of diffusion-weighted MRI by conditional VAE to produce cohort-invariant encodings. Different from other conditional VAE based methods, SH-VAE took spherical harmonics coefficients as input and output.

(2) **stVAE** [132] applied Conditional VAE with Y-Autoencoders (additional classification head in the encoder) and adversarial feature decomposition for single-cell RNA sequencing.

(3) **scAlign** [133] performed harmonisation by learning a duplex mapping of cell sequences between different cohorts in a low dimensional latent space. This mapping enabled the model to estimate a representation of certain samples under data from different cohorts. Besides, it employed the "association learning" method [134] to walk through the embeddings generated by a neural network with data from different cohorts. The association learning enabled the network to extract the embeddings that can capture the essence of the input data, leveraging the lack of annotations in paired synthesis. With these essence embeddings, scAlign applied a decoder to synthesise the harmonised data.

(4) **iMAP** [135] first presented an autoencoder architecture to learn the cohort-invariant features, then used these features to set MNN pairs by the random walk strategy. This autoencoder included one encoder $E$ and two generators $G_1$ and $G_2$, with two inputs (gene expression vectors and cohort labels) and outputs ($G_1$ for generating the cohort variations and $G_2$ for reconstructing the original input), respectively. With the defined MNN pairs, a GAN model was used to produce the cohort-invariant samples.

## 4.4 Invariant feature learning

The invariant feature learning techniques are meant to learn the cohort-invariant features from different cohorts of data, then apply these features for the main task (e.g., segmentation, classification, regression). The concept behind representation learning approaches for harmonisation is that if a sparse dictionary/mapping can be built from data of different cohorts, these learnt representations will not include inter/intra cohort variability.

### 4.4.1 Dictionary learning (DictL)

**Sparse dictionary learning (SDL):** SDL [136] was a representation learning approach that aimed to reduce the complexity of the harmonisation task by decomposing the input data as a linear combination of components. SDL could be applied to identify the cohort-invariant features to reconstruct the raw data from a huge number of random features [137].

**Unsupervised colour representation learning (UCRL):** UCRL [138] first estimated the sparsity parameter based on SPCN, then employed a robust dictionary learning method [139] to acquire the stain colour appearance matrix. By taking the stain centroid estimation as an $L_1$-regularised linear least-squares task, the stain mixing coefficients map of the source data was combined with the colour appearance matrix of the reference data.

### 4.4.2 Autoencoder based methods (AE)

**DESC:** DESC [140] trained a VAE to obtain the cohort-invariant feature embeddings, then iteratively optimised a clustering loss function to group the cohort data. The Louvain clustering [141], which aimed to improve modularity for community detection, was used to initialise the cluster centres.

**BERMUDA:** BERMUDA [142] first applied a graph based clustering to data from different cohorts individually, followed by a method (named MetaNeighbor) to identify similar clusters between cohorts to get the initial unaligned comprehensive dataset. An autoencoder was then built to reconstruct the input data while producing invariant feature embeddings in the low dimensional latent space. These feature embeddings were cohort-invariant and can be used for further analysis.

### 4.4.3 Adversarial learning methods (AdvL)

The adversarial learning methods indicate developing a learning system that focuses on the scanner/protocol invariant features while simultaneously maintaining performance on the main task of interest, thus reducing the cohort bias on predictions. These methods [143-146] were usually composed of an adversarial module for cohort identification, a backbone for feature extraction, and the main task for classification, regression, and/or segmentation.

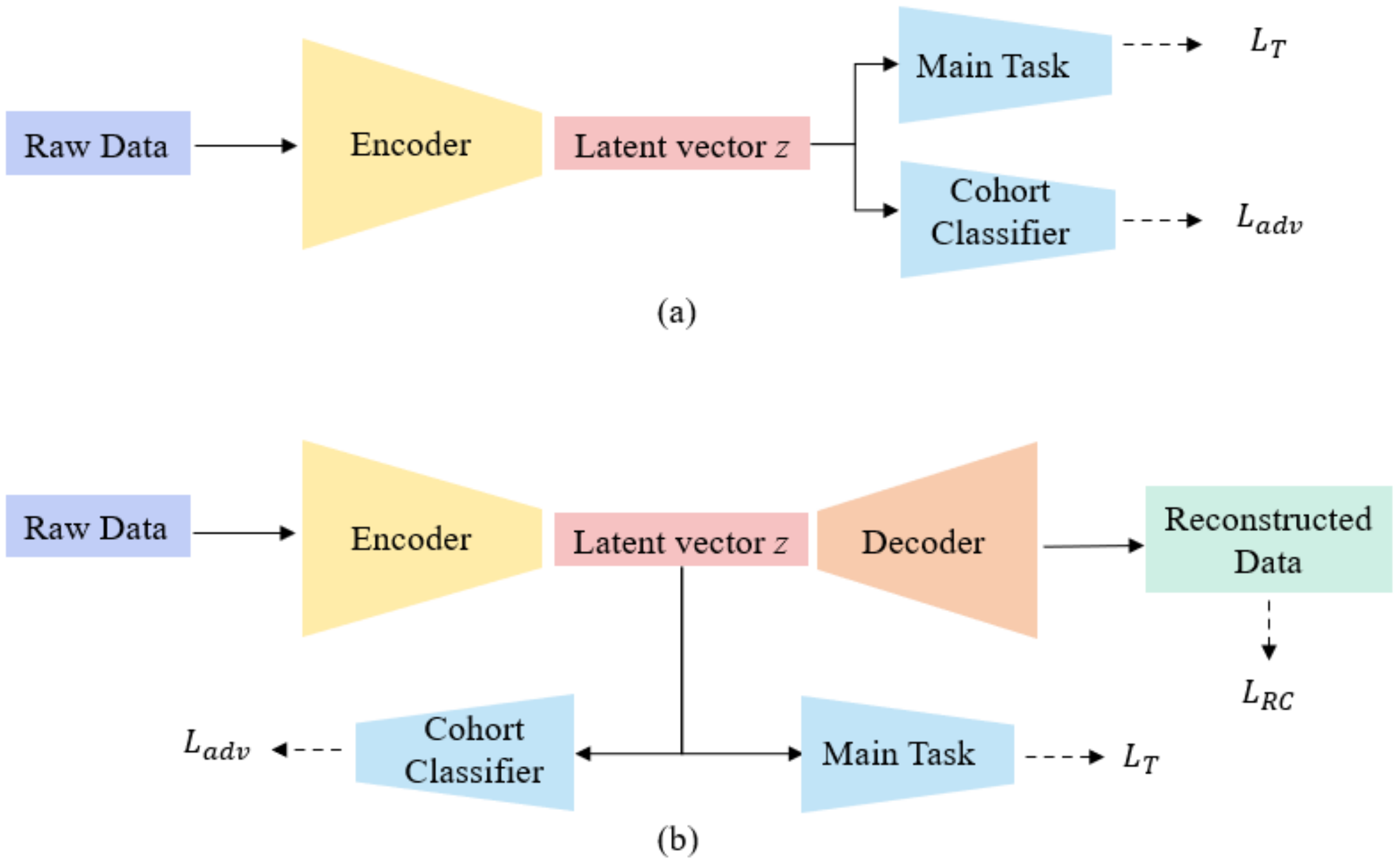

Fig. 5 Illustration of adversarial learning methods.

The adversarial learning methods used for harmonisation mainly had two structures, as shown in Fig. 5 (a) and (b). For methods such as $AD^2AH$ (Attention-guided deep domain adaptation harmonics) [143], DUH (Deep unlearning harmonics) [144], and scDGN (single-cell domain generalisation network) [145], the adversarial module (domain classifier) was designed to assist the encoder to learn the cohort invariant features by maximising the adversarial loss $L_{adv}$ while minimising the loss of the main task $L_T$ (Fig. 5(a)). To acquire the precise feature representation $z$, methods such as NormAE [146] added a decoder to reconstruct the input raw data through minimising the reconstruction loss $L_{RC}$, shown in Fig. 5 (b). By incorporating these optimisation functions, the main task could achieve stable performance when dealing with multi-cohort data.

# 5. Evaluation approaches of the data harmonisation strategies

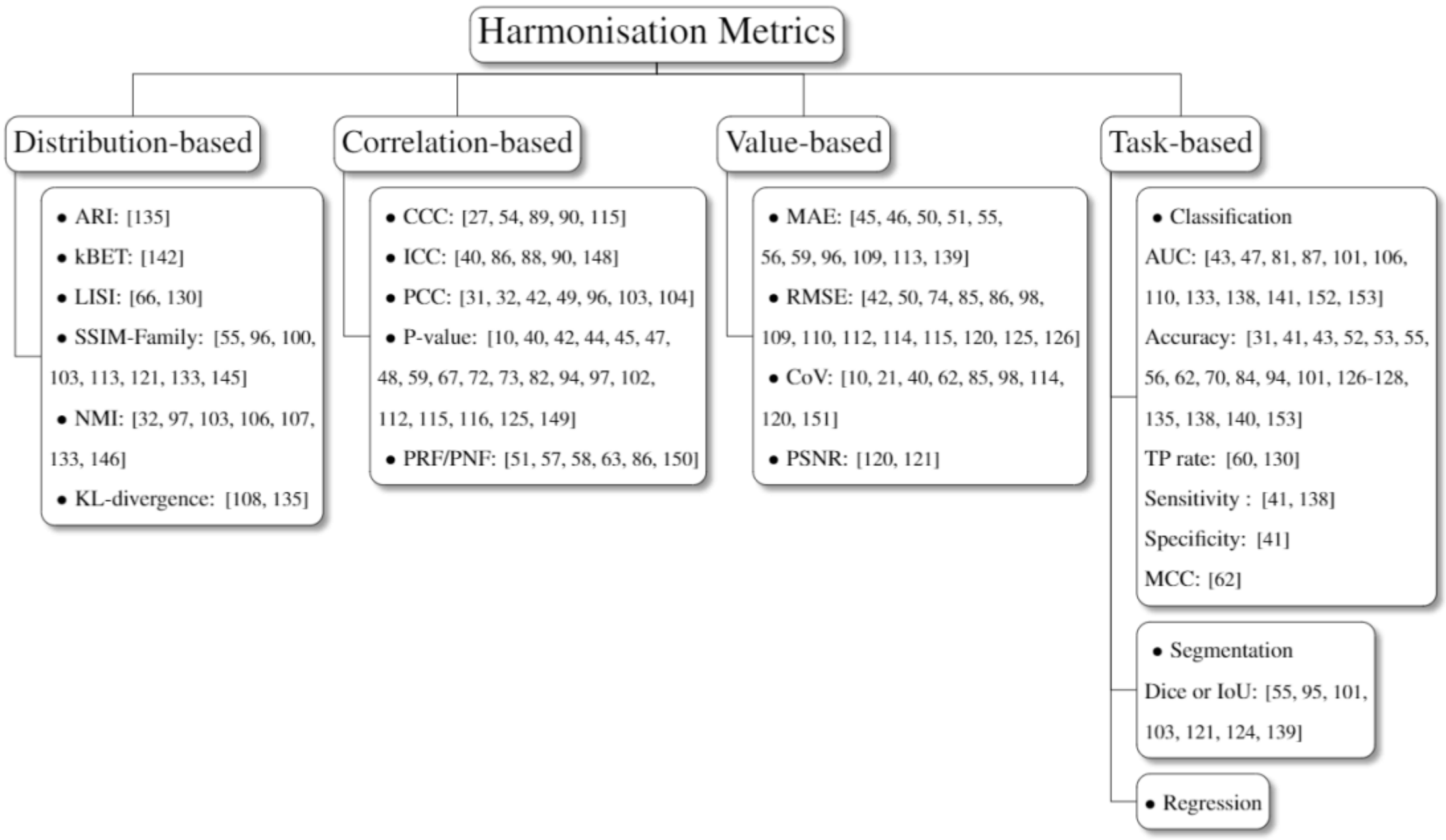

Fig. 6. Taxonomy of harmonisation metrics. The visualization and cohort classification assessment are not presented due to their limited subcategory.

This section explores evaluation approaches for harmonisation performance and divides them into distribution based, correlation based, value based, and task based metrics. Distribution based metrics evaluate the harmonisation performance through assessing the clusters or location-scale parameters among different cohorts. The correlation based and value based metrics assess the variability of data-derived features from different cohorts to test the reproducibility. Besides, cohort classification is also considered as an evaluation method, aiming to demonstrate the harmonisation effect by cohort classification results. Visualisation is the commonly used evaluation approach that can straight visualise datasets before and after harmonisation.

## 5.1 Distribution based evaluation

Distribution based metrics assess the harmonisation performance via calculating the clustering or local-scale parameters. The clustering related metrics include adjusted rand index (ARI), k-Nearest neighbour batch-effect test (kBET), local inverse Simpson's index (LISI). The location-scale related metrics contain structural similarity families, normalised median intensity and KL divergence.

### 5.1.1 Adjusted rand index (ARI)

The adjusted Rand index is the corrected-for-chance version of the Rand index (RI) and can be used for harmonisation evaluations [140]. Given a set of $n$ elements and their predictions $X = \{X_{1,}X_2, \dots, X_i\}\ and\ Y = \{Y_1, Y_2, \dots, Y_j\}$, the RI can be calculated through

$$\mathrm{RI} = \frac{\mathrm{TP} + \mathrm{TN}}{\mathrm{TP} + \mathrm{FP} + \mathrm{FN} + \mathrm{TN}}, \tag{5}$$

where TP is the number of true positives, TN indicates the number of true negatives, FP is the number of false positives and FN is the number of false negatives. The ARI is illustrated as

$$\mathrm{ARI} = \frac{\mathrm{RI} - \mathrm{E[RI]}}{\max[\mathrm{RI}] - \mathrm{E[RI]}}, \tag{6}$$

where E(RI) is the expectation of the RI. It ranges from 0 to 1, and a large ARI indicates the cluster results are similar to the real labels.

5.1.2 k-Nearest neighbour batch-effect test (kBET)

The k-Nearest neighbour batch-effect test was proposed to assess whether the distribution based harmonisation method can remove cohort bias while preserving biological variability [147]. kBET formulates a null hypothesis that the data is 'well mixed'. It employs a $\chi^2$ based test for random fixed-size neighbourhoods to evaluate whether the data is well mixed. The low average rejection rate indicates good harmonisation performance and vice versa. As a result, determining whether the mean rejection rate surpasses a significance level allows the null hypothesis to be rejected for the whole dataset.

5.1.3 Local inverse Simpson's index (LISI)

LISI combines perplexity based neighbourhood construction and the Inverse Simpson's Index (ISI), which is sensitive to local diversity and can be well interpreted [73, 135]. LISI applies the Gaussian kernel based distributions of neighbourhoods via distance based weights and computes the local distributions by fixing perplexity. Meanwhile, it uses the ISI to enhance the interpretation, that is

$$\mathrm{ISI} = \frac{1}{\sum_{n=1}^{N} p(x)}, \tag{7}$$

where *p(x)* is the batch probabilities in local distributions.

5.1.4 Structural Similarity Families

Structural similarity index measure (SSIM) was designed to evaluate the image quality degradation during data transmission by measuring the similarity between two samples [148]. It was initially proposed for grey level images and has been widely applied to evaluate the harmonisation performance in digital pathology [62, 101, 108, 118, 126]. Assume $\mu_i$, $\sigma_i$ as the average and variance of sample $i$, $\sigma_{xy}$ as the covariance between sample $x$ and $y$, and the smooth parameters $c_1, c_2$. The SSIM can be described as

$$\mathrm{SSIM}(x,y) = \frac{(2\mu_x\mu_y + c_1)(2\sigma_{xy} + c_2)}{(\mu_x{}^2 + \mu_y{}^2 + c_1)(\sigma_x{}^2 + \sigma_y{}^2 + c_2)}. \tag{8}$$

To better assess the similarity for colour images, Kolaman et al. [149] proposed quaternion structural similarity (QSSIM) to measure the size and direction of chrominance, luminance and degradation [109]. Feature similarity index (FSIM) utilises phase congruency and gradient magnitude features to evaluate the low-level features of image visual quality [150]. The QSSIM and FSIM were employed in [110, 138] and [105, 110], respectively, to assess the structural preserving conditions after the harmonisation process. Though most methods applied structural similarity related metrics for evaluation, studies have shown their limitations and weaknesses [151]. For instance, it has been reported that these metrics suffer from uniform pooling, distortion and instability, especially when measuring samples with hard edges or low variance regions.

5.1.5 Normalised median intensity (NMI)

Assume the mean of R, G, B values for the i-th pixel within the image $I$ as $A_i$, the NMI for assessing the colour consistency is calculated as

$$\mathrm{NMI}(I) = \frac{\mathrm{Median}(A_i)}{\mathrm{P}_{95}(A_i)}, \tag{9}$$

where the $\mathrm{P}_{95}$ denotes the 95th percentile [39, 102, 108, 111, 112, 138, 152]. The harmonisation strategy is effective when the median and maximum intensity values are close enough. Since the NMI does not consider the consistency of the ROI within the same biopsy set of $S$ images, Maji et al. [113] presented an extension of NMI, named Between-Image colour constancy (BiCC) index, which can be represented by

$$\mathrm{BiCC}(I) = \frac{1}{2(|S|-1)} \times \sum_{J \neq I} \frac{\mathrm{Median}(A_i) + \mathrm{Median}(A_j)}{\mathrm{Max}[\mathrm{Max}(A_i), \mathrm{Max}(A_j)]}, \tag{10}$$

where $i \in \mathrm{ROI}(I)$ and $j \in \mathrm{ROI}(J)$. The value of BiCC ranges from 0 to 1, an efficient harmonisation algorithm for image modality should make the value as high as possible.

5.1.6 Coefficient of variation (COV)

Given the mean $\mu$ and standard deviation $\sigma$, the coefficient of variation (COV) is defined as $\sigma/\mu$, which depicts the degree of variation in respect to the population mean [17, 28, 47, 69, 90, 103, 119, 125, 153]. The Multivariate COV (MCOV) is used to quantify the variability of features between different cohorts, with a lower value indicating better reproducibility [154]. Assume $\mu_x$ and $\mu_y$ as the mean of features extracted from two different cohorts $x$ and $y$, $\sum_{x,y}$ as the covariance matrix, the MCOV is computed via

$$\mathrm{MCOV}(x,y) = \sqrt{\frac{\mu_{x,y}^T * \sum_{x,y} * \mu_{x,y}}{\left(\mu_{x,y}^T * \mu_{x,y}\right)^2}}, \tag{11}$$

5.1.7 Kullback-Leibler divergence

Kullback-Leibler (KL) divergence was proposed to measure how a probability distribution is different from another one. Assume two discrete probability distributions $p$ and $q$ (each with $k$ samples), the KL divergence is given by

$$\mathrm{KL} = \sum p_k log \frac{p_k}{q_k} \tag{12}$$

It was applied as a metric for harmonisation strategies, with 0 indicating identical quantities of information between two distributions [113, 140]. For instance, Li et al. [140] applied KL divergence to evaluate how randomly are samples from different cohorts mixed together within each cluster.

## 5.2 Correlation based evaluation

The measurement of reproducible/nonreproducible is usually given by statistical analysis via calculating the correlation between data-derived features before and after harmonisation, including concordance correlation coefficient, intra-class correlation coefficient etc.

5.2.1 Pearson correlation coefficient (PCC)

Pearson correlation coefficient measures the linear correlation between two groups of variables *X* and *Y*, which is presented as

$$p_{X,Y} = \frac{\mathrm{cov}(X,Y)}{\sigma_X \sigma_Y}, \tag{13}$$

where cov is the covariance and $\sigma$ indicates the standard deviation. PCC ranges from 0 to 1 where 0 denotes there is no correlation between *X* and *Y*, and 1 represents a perfect correlation. The PCC was used as an indicator to assess the similarity between the source and harmonised data [38, 39, 49, 56, 101, 108, 109].

5.2.2 Concordance correlation coefficient (CCC)

The CCC was proposed by Lin et al. [155] that measured the agreement between two variables and was used to assess the reproducibility [34, 61, 94, 95, 120]. Different from PCC that can only assess the correlation between two groups of data, CCC measures how large the gap between two groups of data is. Assume the two variables $X = \{x_1, x_2, \dots, x_n\}$, $Y = \{y_1, y_2, \dots, y_n\}$ and their mean $\bar{x}, \bar{y}$, and variance $s_x^2, s_y^2$, the CCC is given by

$$\rho_c = \frac{2 s_{xy}}{s_x^2 + s_y^2 + (\bar{x} - \bar{y})^2} \tag{14}$$

where $s_{xy} = \frac{1}{N} \sum_{n=1}^{N} (x_n - \bar{x})(y_n - \bar{y})$.

5.2.3 Intra-class correlation coefficient (ICC)

Both PCC and CCC can only assess the correlation between two groups of data, which cannot be implemented on multi-cohorts. The ICC is utilised for data structured as groups instead of those as paired observations, it is usually used to assess the variability within different protocols, different imaging devices, or different sites [47, 91, 93, 95, 156]. It interprets on a scale of [0, 1], with 1 illustrating the perfect agreement and 0 indicating complete randomness. Essentially, the ICC employed for data harmonisation describes the confidence of how similar the variables are in different cohorts, which is the one-way random model that assumes there is no systematic bias [47]. Data from various cohorts are pooled and assessed within or cross operators based on the analysis of variance (ANOVA). The one-way random model can be given from:

$$\mathrm{ICC} = \frac{MS_B - MS_W}{MS_B + (n_G - 1) MS_W} \tag{15}$$

where $MS_B$ is the mean square between groups, $MS_W$ is the mean square within groups and $n_G$ indicates the average group size.

5.2.4 P-value

Some studies evaluate the harmonisation effectiveness through computing the P-value given by paired hypothesis tests [17, 47, 49, 51, 52, 54, 55, 66, 74, 79, 80, 87, 99, 102, 107, 117, 120, 121, 130, 157]. This statistical analysis is often conducted based on the paired region of interests before-and-after harmonisation. In particular, there are significant differences (corresponding to p-value < 0.05) between the data of different cohorts before and after harmonisation, and vice versa. For instance, Fortin et al. [58] analysed the number of voxels that are significantly related to cohorts, e.g., a voxel is counted when the p-value calculated is less than 0.05.

### 5.2.5 Percentage of reproducible/nonreproducible features (PRF/PNF)

The percentage of nonreproducible features was treated as an evaluation metric [58, 64, 65, 70, 91, 158], e.g., Mahon et al. [158] compared the percentage of significantly different features before and after conducting ComBat harmonisation. On the contrary, the percentage of reproducible features was also considered as an evaluation metric of harmonisation performance [61].

## 5.3 Value based evaluation

The value based evaluation mainly assesses the intensity differences between the data or data-derived features before and after harmonisation. This usually requires a "ground truth" that can ideally reflect harmonisation results, a low value of intensity differences indicates good harmonisation performance.

### 5.3.1 Mean absolute error (MAE)

The average absolute error of features (textual and clinical features) can be used to reflect harmonisation effects [52, 53, 57, 58, 62, 63, 66, 101, 114, 118, 144], this usually requires the extraction of certain ROIs from the data before and after harmonisation. For instance, Wachinger et al. [63] evaluated the MAE in age prediction on the raw dataset and ComBat-harmonised dataset to illustrate the effectiveness of ComBat. Dewey et al. compared the MAE between the synthesised and raw images to demonstrate the harmonisation performance.

### 5.3.2 Root-mean square error (RMSE)

Many researchers measured the RMSE between the harmonised samples and the ground truth targets to assess the replicability [49, 57, 81, 90, 91, 103, 114, 115, 117, 119, 120, 125, 130, 131]. For instance, Moyer et al. employed RMSE and mean absolute error to assess the harmonisation performance between synthesised diffusion MRI and the ground truth. However, this metric requires paired datasets which is a heavy burden for digital healthcare research.

### 5.3.3 Peak signal to noise ratio (PSNR)

PSNR illustrates the ratio between the maximum power of a signal and the power of noise that influences the integrity of its representation. Consider two groups of variables $X$ and $Y$, the PSNR between $X$ and $Y$ can be given as

$$\mathrm{PSNR} = 20 \times log_{10}\left(\frac{\mathrm{MAX_i}}{\sqrt{\mathrm{MSE}(X,Y)}}\right), \quad (16)$$

where $\mathrm{MAX_i}$ denotes the maximum possible value of the image (255 for images) and MSE is the mean squared error. It is commonly used in image denoising tasks as an evaluation metric. Some researchers applied this indicator to measure the quality of the synthesised images during the harmonisation [125, 126].

## 5.4 Main task based performance evaluation

Many studies demonstrated its effectiveness by comparing the performance of the main tasks before and after harmonisation methods. Although it is a result-oriented evaluation method and may be affected by the random initialisation of the main-task models (machine learning models), it can prove the effectiveness of the harmonisation method to some extent. The main tasks involved in harmonisation evaluation mainly include regression [57, 68, 91], segmentation, and classification [84]. The assessment is usually done by the Dice coefficient (Dice) or the intersection over union (IoU) [62, 100, 106, 108, 126, 129, 144]. On the contrary, the classification tasks are variously evaluated, e.g., using area under the receiver operating characteristics curve (AUC) [50, 54, 86, 92, 106, 111, 115, 138, 143, 146, 154, 159], accuracy [38, 48, 50, 59, 60, 62, 63, 69, 77, 89, 99, 106, 131-133, 140, 143, 145, 159], true positive rate [67, 135], sensitivity [48, 143], specificity [48] and Matthews correlation coefficient (MCC) [69]. Note that MCC is a balanced measurement for the binary classification tasks, with comprehensive evaluations of TP, TN, FP, and FN, therefore it is divided into the main task based performance evaluation.

## 5.5 Cohort classification

Different from comparing the variety of data or data-derived features before and after harmonisation, some studies reported the effectiveness of harmonisation strategies through adopting cohort classification [49, 63, 94]. The core idea of this metric is that the cohort should be more difficult to identify when an effective harmonisation strategy is employed. For instance, Wachinger et al. [63] compared the accuracy of cohort identification of the raw dataset, dataset applied with z-score normalisation, linear model and ComBat, respectively. The worst results were gained by ComBat, which indicates the best harmonisation ability since the classifier cannot well identify the cohorts after harmonisation.

## 5.6 Visualisation

Visualisation refers to the techniques that can picture the data distribution from low dimensional feature space or sample intensities. Approaches for harmonisation assessment that visualising the data distribution in latent space including principal component analysis (PCA) [56, 57, 59, 68, 70, 81, 83, 146], uniform manifold approximation and projection (UMAP) [38, 50, 77, 133, 135, 142, 160], and t-distributed stochastic neighbour embedding (t-SNE) [54, 74, 75, 79, 83, 121, 132, 133, 143, 145]. These approaches convert different high-dimensional data into low-dimensional data and plot them into the same feature space. The harmonisation is well performed if the visualisation of data distribution is mixed instead of assembling as different clusters. In addition to visualising data distributions, some researchers also plot the intensities or location-and-scales (mean and variance) of each sample before and after harmonisation to assess the performance [46, 83, 86, 97, 98, 100, 103, 106, 107, 125].

# 6. Applications of computational data harmonisation

Data harmonisation has been widely adopted in various fields of digital healthcare, including the manual harmonisation of tabular data, computational data harmonisation of the gene, radiomics and pathological data. Though there have been many efforts on removing the artefacts of time series signal data (such as EEG, ECG) [161], these works mainly focus on the removal of noises caused by biological variances. For instance, researchers employed filtering [162] and wavelet transform [163] approaches to remove the ocular, muscle and cardiac artefacts. However, these studies barely pay attention to the device/site variances, indicating the lack of harmonisation studies to these time series signal data. This section illustrates the application of computational data harmonisations in gene expression, radiomics and pathology.

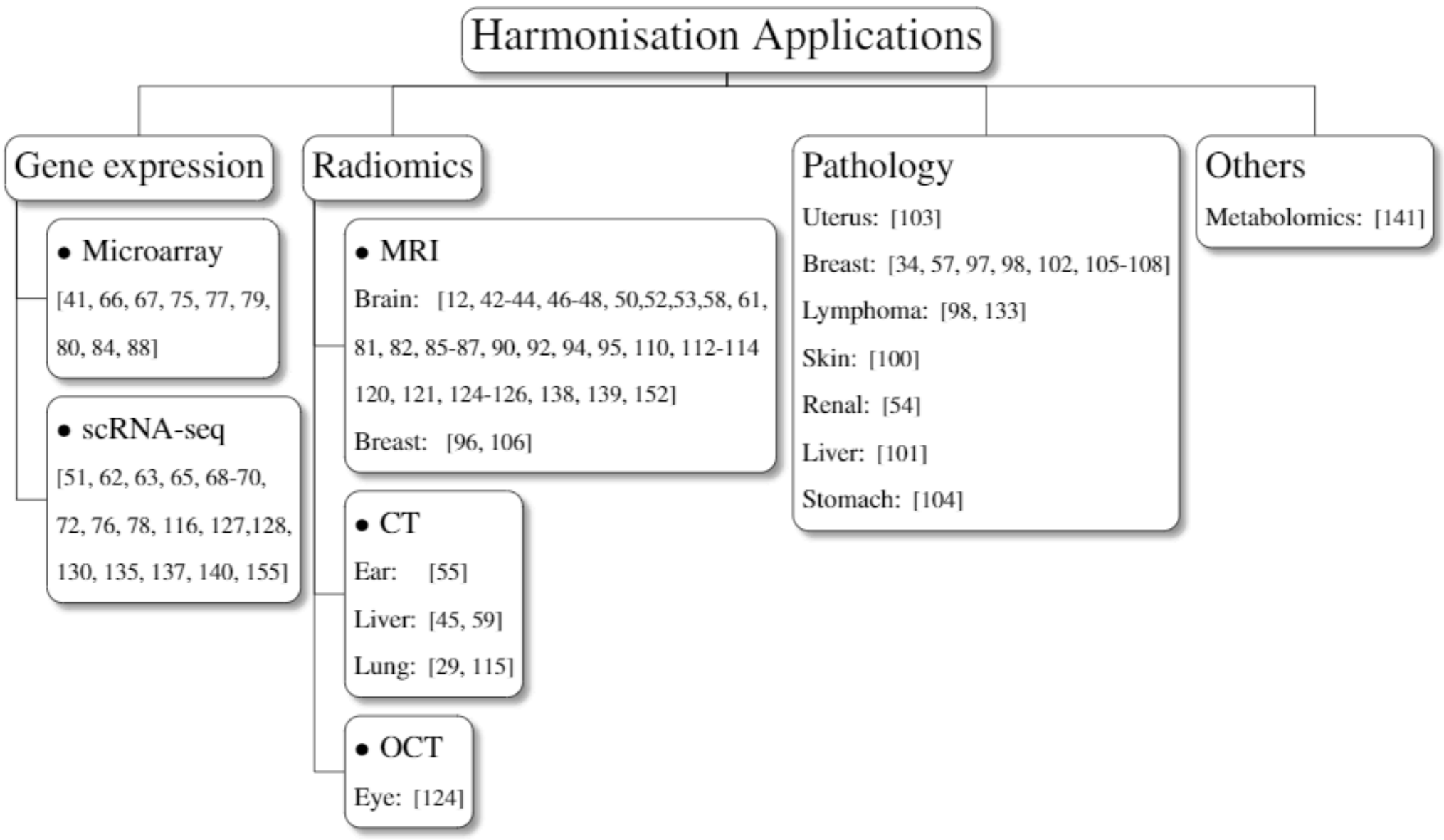

Fig. 7 Taxonomy of applications that involved computational data harmonisation strategies.

## 6.1 Gene expression analysis

The process of generating a functional gene product from the information within a gene is referred to as gene expression, which is one of the major research areas in biomedical research. The traditional approach of gene expression analysis is microarray technology, which relies on comprehensive chemical reactions to convert RNA to cDNA. The latest gene expression analysis method is single-cell RNA sequencing (scRNA-Seq). It isolates the single cells and RNA for transcription, library generation and sequencing, using the new generation sequencing (NGS) techniques. Unfortunately, due to the various NGS platforms and experimental environments (pH, temperature), both microarray technology and scRNA-sequencing are highly affected by cohort bias. Therefore, computational data harmonisation is widely used in microarray [46, 71, 72, 80, 82, 84, 85, 89, 93] and scRNA-Seq [56, 67, 68, 70, 73-75, 77, 81, 83, 121, 132, 133, 135, 140, 142, 145, 160] to remove the cohort bias.

## 6.2 Radiomics analysis

Radiomics refers to the extraction and analysis of a large number of quantitative image features from medical images obtained by CT, PET, MRI, and other imaging modalities [164]. In addition to the studies on phantoms [61, 65, 94], research refer to MRI mainly focus on the brain [17, 47-49, 51-53, 55, 57, 58, 63, 66, 86, 87, 90-92, 95, 97, 99, 100, 115, 117-119, 125, 126, 129-131, 143, 144, 157], breast [101, 111], while those refer to CT focus on ear [60], liver [50, 64] and lung [34, 120]. In addition to CT, PET and MRI, optical coherence tomography also suffers severe scanner variability, which can be eased by computational data harmonisation approaches [129].

## 6.3 Pathology analysis

Most harmonisation strategies (also named stain/colour normalisation) in pathology aimed to address the stain variance. These studies mainly focused on uterus [108], breast [39, 62, 102, 103, 107, 110-113], lymphoma [103, 138], skin [105], liver [106], renal [59], and stomach [109].

## 6.4 Other modalities

Metabolomics is an omics technology that monitors and discovers metabolic changes in people in relation to illness state or in reaction to a medical or external intervention using current analytical instruments and pattern recognition algorithms. The nonlinear cohort bias during the liquid chromatography−mass spectrometry can be removed by computational methods [146].

# 7. Meta-analysis

In this review, the methodologies and metrics were grouped based on different ideas or theories, and the meta-analysis was conducted and reported in three areas/modalities. The reason why the results were explored and discussed through different modalities (gene, radiomics, and pathology) is that these data have different properties. For instance, data of gene analysis are expression matrices, data of radiomics are high-dimensional volume array (grayscale image per slice), and data of pathology are colour images with huge sizes.

## 7.1 Meta-analysis

**Data properties and study trends.** The number of studies and data properties for harmonisation approaches from 2000 to 2021 is demonstrated in the top left in Fig. 8, with the percentage of studies that was conducted on the public dataset. The public data can be acquired through open source websites or archives while the in-house data cannot be acquired. There has been a dramatic increase in the number of harmonisation studies since 2019, indicating an urgent need to conduct large-scale studies and data harmonisation strategies. In addition, we demonstrate the number of harmonisation studies on different sub-modalities in recent years (Microarray and sRNA-seq for gene expression studies, CT and MRI for radiomic studies, and Pathology). In terms of gene expression, the harmonisation approaches for microarrays were mainly presented before 2015, while that for sRNA-seq have become the latest topic in recent years. As for radiomics studies, researchers have realised the importance to improve the reproducibility of radiomics features, especially in MRI. Data harmonisation for digital pathology has been noted in decades, while it receives more attention in recent years.

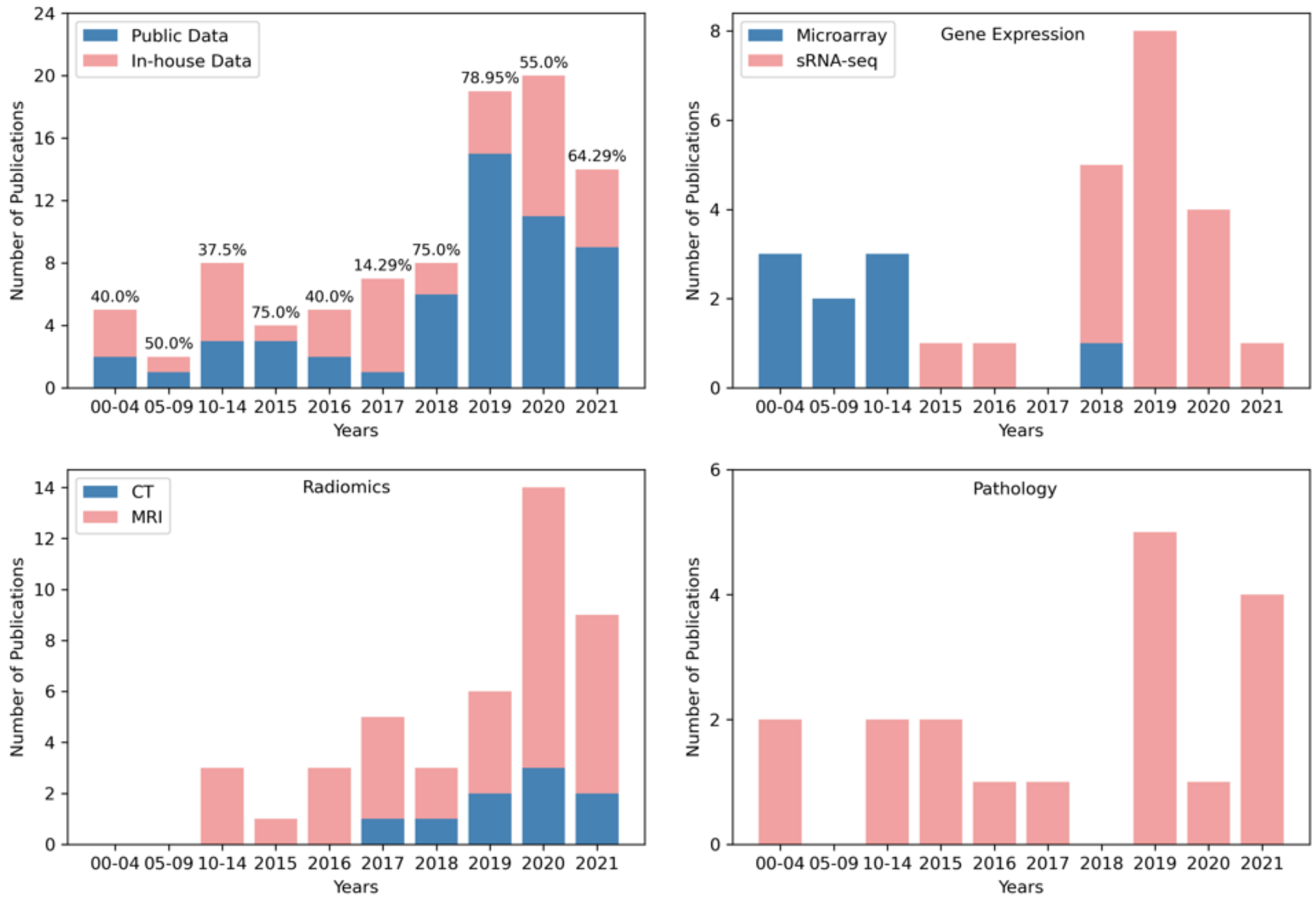

Fig. 8 Number of publications and years in terms of data properties and modalities. The public data is the open source data that can be acquired, the in-house data is not available from the internet. The percentage in the top left subfigure is the ratio of studies that were conducted on the public dataset.

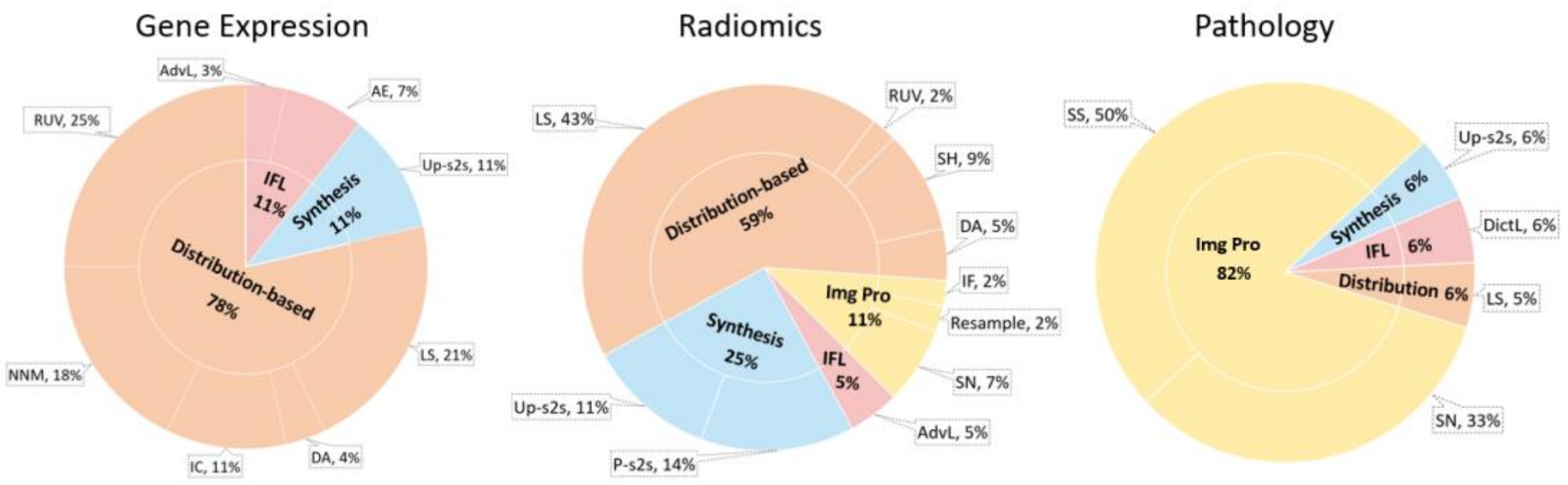

Fig. 9 Harmonisation strategies in terms of different modalities. 'IFL' indicates invariant feature learning approaches, "Img Pro" refers to image processing approaches. The percentage of sub-methods is annotated with the abbreviations of sub-methods in each pie chart.

**Strategies and modalities.** Due to the diversity of biomedical data modalities, the relationship of different strategies and modalities was explored. As shown in Fig. 9, the distribution based methods were commonly applied in gene expression and radiomics studies, which account for 79% and 59% of the employed approaches, respectively; however, only a few of them (5%) were employed in digital pathology. The empirical Bayes methods dominated the distribution based methods because of their generalisation ability and robustness, while the RUV and SH were more commonly used in specific fields. The image processing approaches were mainly used in digital pathology, employing standardisation/normalisation and stain separation ideas to merge the multi-cohort data. Unlike the distribution based and image processing based methods, invariant feature learning and synthesis were found to be applicable in all three modalities recently, dominated by deep learning based algorithms.

**Evaluation metric.** The evaluation metric is another crucial aspect when developing computational data harmonisation strategies. It describes the performance of harmonisation methods via analysing the distribution, correlation and values between the source and target cohorts. Among all evaluation metrics, visualisation was the most commonly used method to present harmonisation effects, followed by evaluating the main tasks (Fig. 10). Some studies tried to evaluate via classifying the cohorts, but this may have a limitation since the inability to distinguish cohorts does not mean that all data is well harmonised. Overall, even there are many options for the assessment of harmonisation strategies, there still exist some barriers to implementing harmonisation assessment in clinical flows. The evaluation can only be acquired when (1) there are paired datasets (which is inapplicable in real clinical settings); (2) there is a certain machine learning-based module for performance comparison (demands for well-trained computational modules); (3) there are clinicians for visual assessment (subjective and time-consuming); (4) there are predefined regions of interest (demands for manual annotation or computational modules). Moreover, data harmonization is of utmost importance for a seamless federation of models (i.e. a naïve federation approach in which no additional algorithms are needed to cope with incoherencies in local datasets). Therefore, to which extent should local datasets be harmonized not to destroy locally contextual particularities of the data that positively contribute to the local generalization of the models.

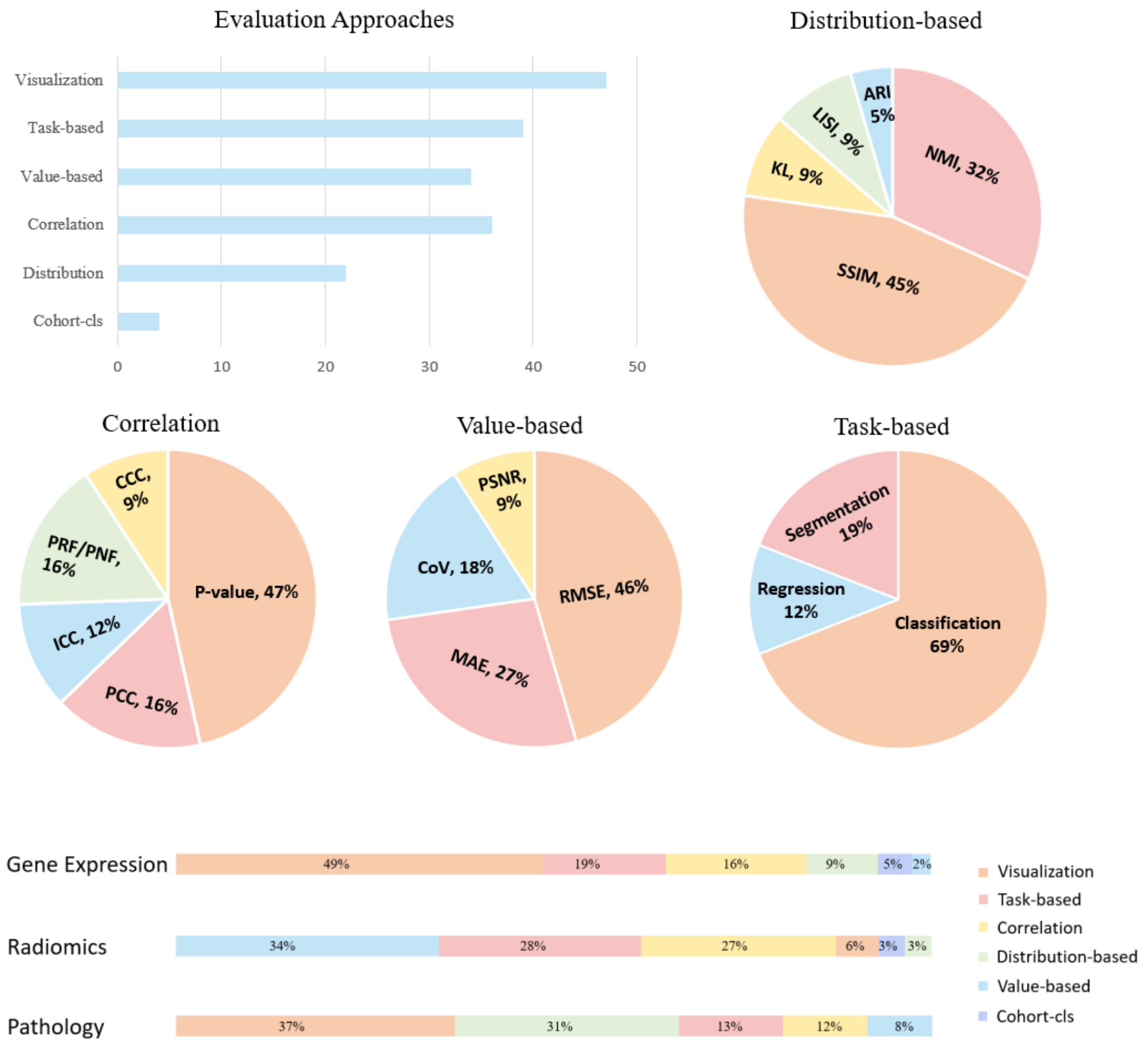

Fig. 10. Evaluation metrics in terms of different modalities.

In addition to reporting the utilisation of different metrics, details of evaluation metrics in terms of different modalities were presented in Fig. 10. Visualisation is preferred for gene expression and digital pathology, including the visualisation of data distribution (e.g., using UMAP, t-SNE) and samples before and after harmonisation. Many pathology studies applied distribution based metrics, such as structure similarity and normalised median intensity. The task based evaluation was also considered reliable, which accounts for 19%, 28% and 13% of all evaluation metrics.

Table 4. Data scale (image size) in previous studies.

| | Small | Middle | Large | N/A |
|---|---|---|---|---|
| Radiomics | 9.1% (6) | 18.2% (12) | 3.0% (2) | 69.7% (46) |
| Pathology | 5.0% (1) | 15.0% (3) | 60.0% (12) | 20% (4) |

* The small, middle, large image sizes are defined as $\epsilon \in [0, 256)$, $\epsilon \in [256, 512)$, $\epsilon > [512, \infty)$, respectively, N/A indicates there is no report of image size.

**Data scale-images.** The scale of samples in radiomics and pathology is closely related to image resolution and qualities. We analysed the variable $\epsilon = \sqrt{w * h}$ of studies that involved images with width $w$ and height $h$. Most radiomics studies were performed with 256 pixels while some were conducted with 512 and 128 pixels, this was because of the shortage of GPU or RAMs especially when dealing with 3D or multi-slice datasets. Moreover, it is of note that 69.7% of radiomics studies did not report the image size, while that proportion of pathological studies was 20%. For pathological images, most studies were conducted with large scale images ($\epsilon > 500$), since the image processing algorithm performs better results when considering the larger field of views.

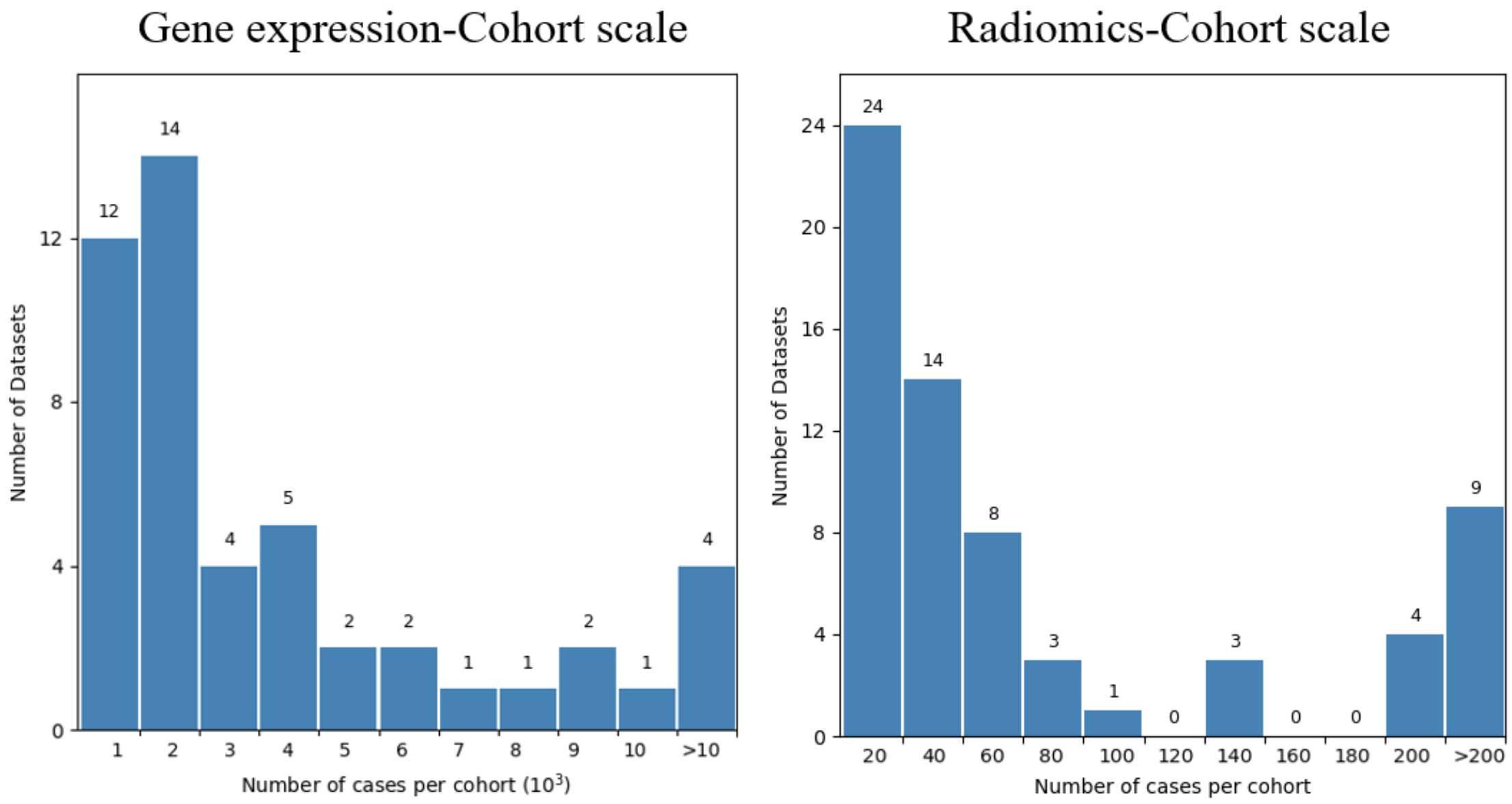

Fig. 11. Scales of cohorts in gene expression and radiomics studies.

**Data scale-cohort.** In addition to the sample (image) scales, the scale of cohorts is also important for data harmonisation approaches. For gene expression, more than half of the studies were performed on large datasets (number of cases per cohort > 5000). Most radiomic studies were conducted on small datasets, with the number of cases (scans) per cohort < 200.

## 7.2 Research directions

This section first presents research tracks for different kinds of data harmonisation approaches based on their limitations, respectively, and states the common restrictions of previous studies.

**Directions for distribution based methods.** The distribution based methods mainly map the source data to the target one through the estimation of the cohort variances. This leads to issues that (1) most distribution based methods were conducted based on refined feature vectors that required prior knowledge of the region of interest. This prior knowledge conflicts with the original purpose that harmonisation approaches are proposed to process multicentre datasets to build robustness and precise computational tools because the region of interest cannot be well predicted by models trained without data harmonisation; (2) although studies have proved that some distribution based methods (such as ComBat) can remove cohort bias while preserving the differences between radiomics features on phantoms, all these methods cannot be well performed to images or high dimensional signals, due to demanding computational complexity; (3) the data harmonisation needs to be performed to entire datasets again when new data are added; (4) some approaches are pairwise (e.g., XPN, DWD, CCA, MNN, Seurat), leading to a complex training procedure (repeated training) when they are implemented to multicentre datasets (number of cohorts > 2). In particular, the first cohort will be considered as the target cohort to correct samples in the second one, and these corrected samples are then added to the first cohort [140].

To overcome these problems, researchers may (1) focus on the harmonisation of raw datasets, instead of data-derived features; (2) develop highly efficient data harmonisation approaches that can deal with a large amount of data; (3) enhance the robustness of data harmonisation strategies; (4) develop methods that can simultaneously harmonise multicentre datasets; and (5) avoid using pair-wise samples for algorithms development.

**Directions for image processing based methods.** Image processing based methods can harmonise the image data without complex procedures. However, these methods also have limitations as (1) some of them (such as stain separation) can be only performed to specific fields; (2) some (image filtering) methods heavily rely on empirical settings, such as filtering kernel sizes and kernel types, which are less efficient and hard to reproduce; and (3) some may lose the information during non-linear transforms. To address these issues, researchers should pay attention to general data harmonisation approaches that do not heavily rely on empirical settings.

**Directions for synthesis.** Although deep learning based synthesis solutions have advanced rapidly and achieved significant performance, these methods still suffer from poor reproducibility and generalisability. The obvious limitations are (1) most synthesis methods were built based on the existing multicentre datasets, which lack evaluations on new datasets; (2) the GAN based models are prone to instability and may hallucinate or introduce unrealistic changes; and (3) training a GAN based model requires a large amount of training data for all cohorts, which may be not feasible for clinical studies. To overcome these drawbacks, researchers should (1) report the data harmonisation performance on new datasets that are not involved during model development; (2) enhance the stability of data synthesis; (3) build data harmonisation strategies using less training data.

**Directions for invariant feature learning.** Invariant feature learning can reduce the disadvantages of synthesis approaches by learning how to extract cohort-invariant features from datasets, but it still faces some challenges. For instance, it can only extract invariant features for analysis while cannot obtain harmonised data. Therefore, future studies should focus on how to generate the harmonised data using extracted invariant features.

**Explainable AI and harmonisation studies.** Another research niche that still remains uncharted in the literature related to data harmonisation is the use of explainable Artificial Intelligence (XAI) methods [165] to identify possible reasons for incoherent data representations. We envision that XAI approaches can be exploited to gain insight on which visual artefacts are present in data instances that imprint a bias on the predicted outcome of a data-based model. This insight can be then analysed to decide whether the rooting cause of such biasing artefacts correspond to insufficient harmonisation of medical data before the learning phase. Furthermore, out of distribution examples can be also detected by virtue of local explanatory techniques (e.g., those capable of discerning which parts of the input to a model are pushing their output towards one class or another), which upon inspection can be attributed to other exogenous phenomena that can relate to data harmonisation, such as a possible miscalibration of the medical equipment or a change in the protocols capturing the data themselves. On the other hand, better harmonisation has benefits to XAI, since all the data are harmonised into the same standard and no cohort biases would be introduced to the XAI system [166, 167]. All in all, we foresee an interesting research cross-fertilization at the crossroads between harmonisation and XAI.

**Limitations for methodology design**. Most studies for data harmonisation did not follow a stepwise design methodology, which cannot be reproduced easily by third parties. For instance, as shown in Table 4, more than half of the radiomic studies did not report the image scale. Moreover, the different definitions of ‘reproducible’ in previous studies and various evaluation metrics greatly hinder the method comparison for further research.

# 8. Checklist and guidance

To address the issues of methodology design, we presented a Checklist for Computational Data Harmonisation in Digital Healthcare (CHECDHA) to enhance the reproducibility and methodological principle, inspired by the Checklist for Artificial Intelligence in Medical Imaging (CLAIM). Furthermore, the guidance on how to choose data harmonisation strategies is also presented in this section.

## 8.1 Checklist criteria

The proposed checklist clarifies the common practice for data harmonisation through data, model, evaluation, result, and discussion, shown in Table 5.

Table 5. Checklist for Computational Data Harmonisation in Digital Healthcare (CHECDHA) criteria.

| Category | | Item | Explanation | Example |
|---|---|---|---|---|
| Motivation | | Background | The application field of the dataset(s) | Information fusion of DW-MRI data from different scanners |
| | | Importance | Why this study is conducted, how important it is | Dramatically increase the statistical power and sensitivity of clinical studies |
| Data | Common | Dataset | What the dataset(s) is (are), how it is (they are) collected (details of acquisition protocols, entry and exit criteria)<br>How many categories, cohorts, subjects, and cases are included in the studies | $m$ healthy subjects under $n$ protocols ($m \times n$ cases, $n$ cohorts)<br>Protocol 1: …<br>Protocol 2: … |
| | | Property | Whether the dataset(s) is (are) in-house or public, provide the access link if appropriate | Public/In-house |
| | | Pre-processing | How the dataset is pre-processed | Z-score normalisation |
| | | Ground truth | What the ground truth is and how it is generated | Cohort $x$ under protocol $i$ |
| | | Partition | For machine learning, how the dataset is partitioned into training, validation, and testing subsets in terms of the number of samples, patients | 7:2:1 for training, validation and test |
| | | Augmentation | For machine learning, how the dataset is augmented | Randomized flip, rotation |
| | Specific | MRI sequence | What the MRI sequence is | Diffusion-weighted |
| | | Region | Which region(s) of the body or the subject in the dataset is (are) covered | Brain |
| | | Slice size | What the sizes of each slice are | 512×512 |
| | | Pixel/Voxel size | What the physical length of a pixel/voxel is | 0.25mm/ $1mm^3$ |
| | | WSI size | What the sizes of the whole slide images are | 12000×30000 |
| | | Patch size | What the extracted image patches are | 256×256 |
| | | mmp | What the microns per pixel in the level-0 scan are | - |
| Model | | Workflow | What the procedures of train and inference are, illustrated by the flow chart(s) if appropriate. | - |
| | | Learning approaches | What the learning method is. e.g., supervised learning, un/semi-supervised learning | Semi-supervised learning |
| | | Architecture | What the structure of the proposed neural network is, if appropriate | nnUNet |
| | | Task | The description of main tasks conducted on harmonised datasets, e.g., lesion segmentation/classification. | Tumour Segmentation |
| | | Input domain | What the input modality of the proposed method is | 3-D images / 2D feature vectors |
| | | Input size | The input sizes of the model | $n \times w \times h \times c$ |
| | | Loss | What the optimisation functions are during the training. | Dice and cross-entropy loss |
| | | Open-source | Whether the source code is available or not, provide the link if appropriate. | Open-source code www.github.com... |
| | | Platform | The learning library used to build the model | TensorFlow 2.5.0 |
| Evaluation | | Statistical Analysis | What the evaluation methods of statistical analysis are | ANOVA-test |
| | | Metric | What indicators are used to evaluate harmonisation performance, e.g., the ratio of the reproducible features, coefficient of variation, Pearson correlation coefficient. | Intra-class correlation coefficient (>0.9 is considered reproducible) |
| | | Comparison | What existing approaches are used to compare the performance of the proposed method | stVAE |
| | | Visualisation | What approaches are used to visualise the data distribution before and after harmonisation strategies | t-SNE/UMAP/PCA |
| Result | | Result | What the quantitative values of evaluation metrics are. | - |
| | | Time-consuming | The computational time of the proposed method and the comparisons. | 30s per case |
| Discussion | | Novelty | What the innovation of the proposed method is. | - |
| | | Strength | The importance/significance of the issue addressed by the proposed method. | - |
| | | Limitation | What remained and unsolved issues are. | - |
| | | Future works | Whether there will be potential studies in the future. | - |

The proposed CHECDHA checklist can greatly standardise the process of data harmonisation studies, which comprehensively describe the motivation, data, data harmonisation strategy, evaluation and conclusions. Start with a clear motivation (Fig. 12), researchers should first emphasise the importance of performing data harmonisation in a certain field. Then, the compositions of datasets should be illustrated in detail, including the common and specific attributes shown in the checklist. When introducing methodologies, the authors should clearly state their ideas and implementation details (input domain, architecture, input size, development platform, etc.). During the evaluation, researchers should assess the reproducibility using new/independent data or data-derived features before and after data harmonisation by appropriate metrics. Meanwhile, the data harmonisation performance of previous approaches should be considered as comparisons to reflect the advantages of the proposed method. At last, the novelty, strength, limitations and future works should be given in the discussion and conclusion sections.

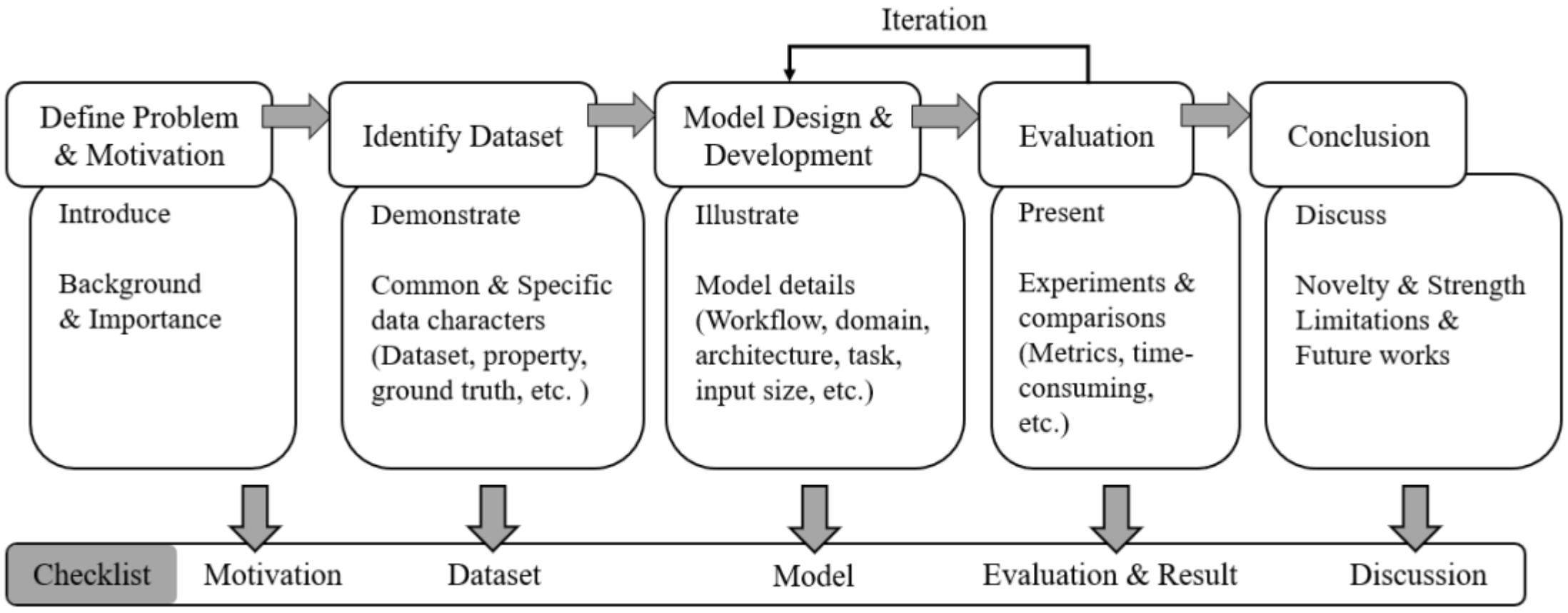

Fig. 12 Workflow of conducting data harmonisation studies guided by the checklist.

## 8.2 Guidance of data harmonisation strategies and metrics

Studies have shown that implementing inaccurate data harmonisation strategies may lead to significant bias, which results in more inaccurate predictions [168]. To guide the method selection, a flowchart presenting possible ways of data harmonisation is presented in Fig. 13. As the flowchart illustrates, the distribution based methods can be well performed on refined features or gene matrices. For high dimensional images, image processing methods are recommended when a high-performance GPU is not available. The deep learning based methods (including invariant feature learning and synthesis) can be applied to all kinds of modalities, while it requires sufficient training samples. The invariant feature learning methods are recommended when the main task can be integrated with the training process, since the synthesis may introduce unrealistic artefacts to the data.

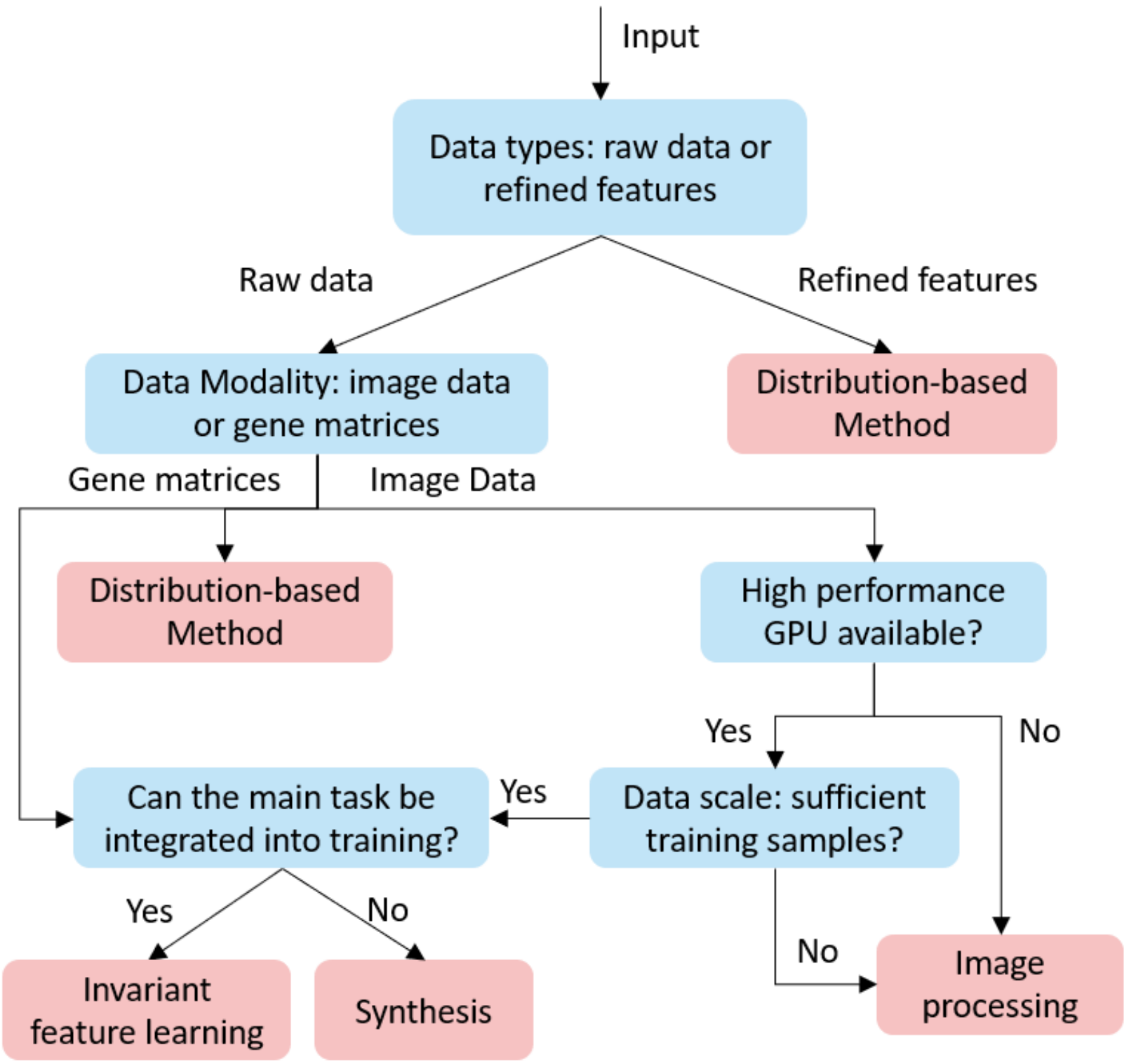

Fig. 13 Flowchart of how to select data harmonisation strategies.

For evaluation, the selection of metrics can directly affect whether the results are reliable or not. Here we summarise and recommend data harmonisation metrics based on different conditions in Fig. 14. Visualisation is the most intuitional way to analyse data harmonisation results, which can be implemented by visualising the raw data with t-SNE/UMAP/PCA or visualising the data harmonised raw data. Main task based evaluation can directly illustrate the effectiveness of the data harmonisation strategies, by comparing the main task performance on data before and after the data harmonisation. If the harmonised ground truth is not available, one can use distribution based metrics to assess the degree of sample mixture (although this may require the cohort label). When the harmonised ground truth can be acquired, the value based or correlation based metrics can precisely present the data harmonisation performance.

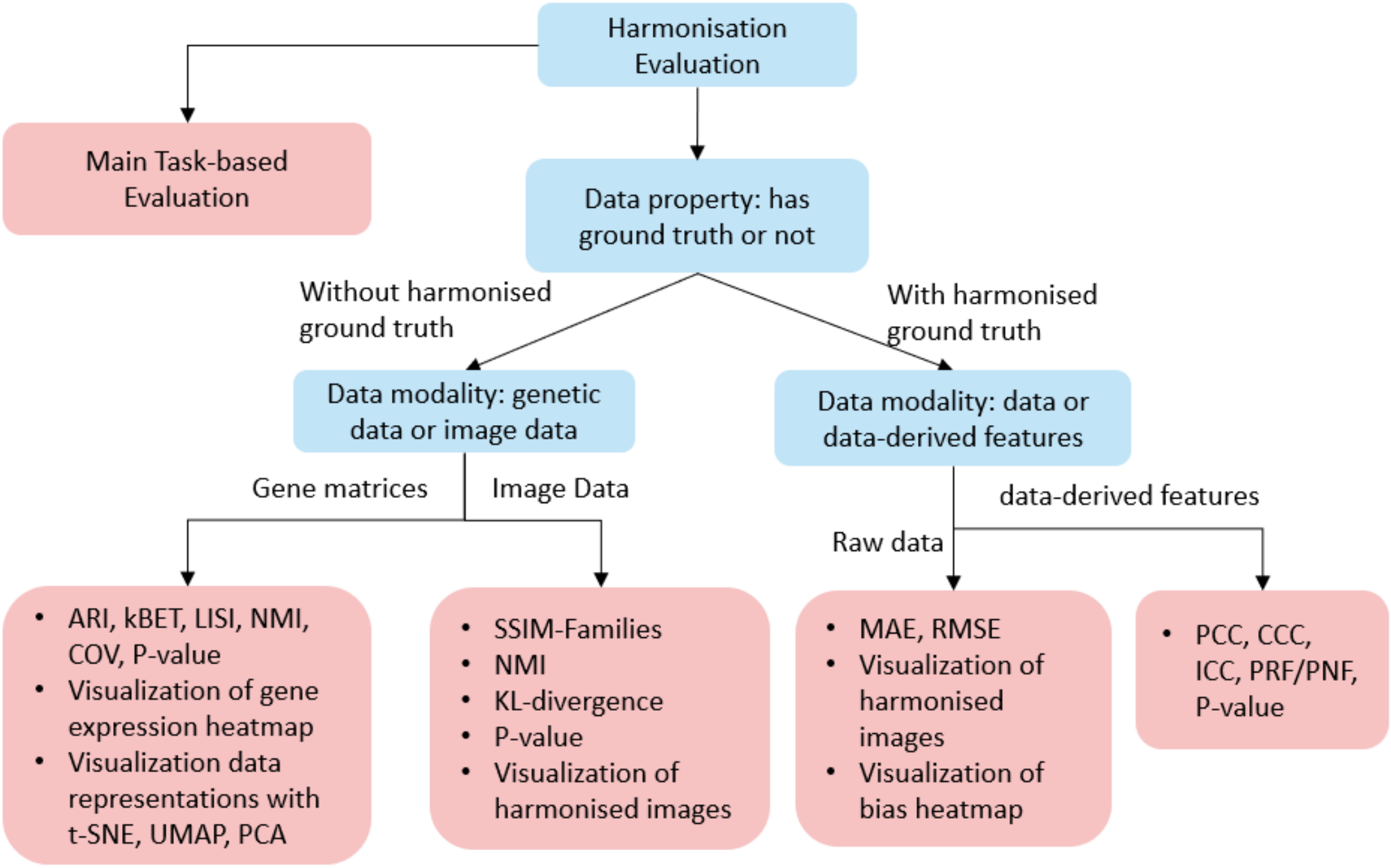

Fig. 14 Flowchart of how to select harmonisation metrics.

## 9. Conclusion

Computational data harmonisation has been proposed for digital healthcare research studies in decades. However, bridging basic science research models and data fusion into multicentre, multimodal and multi-scanner medical practice and clinical trials can be challenging unless data harmonisation can be performed effectively. Furthermore, transfer/federated/multitask learning and other areas wherein knowledge is exchanged among models only work under ideal conditions, whenever the distribution shift is not large enough for the exchange knowledge to remain coherent across models/centres working over different data sources. Otherwise, data harmonisation is needed. Unfortunately, it is unclear which approaches and metrics should be employed when dealing with multimodal datasets. Moreover, there lacks a 'standardised' stepwise design methodology, which leads to poor reproducibility of the existing studies.

To overcome these issues, this paper summarises and categorises the existing data harmonisation strategies and metrics based on different theories, and subsequently presents the CHECDHA criteria. The proposed CHECDHA criteria help researchers to conduct data harmonisation studies in a standardised format, which can greatly advance academic reproducibility and development. Moreover, data harmonisation approaches and evaluation metrics in terms of three modalities are summarised to help researchers to select appropriate strategies (Fig. 7 and Fig. 8). In addition to summarising the methodologies, guidance of method and metrics selection (Fig. 11 and Fig. 12) is also provided according to the different conditions. Last but not least, limitations and directions of different methods are illustrated for future works.

Data harmonisation, an important process in large multicentre studies, has drawn more and more attention in computational biomedical research. It can be well adapted to a federated learning system to promote the development of computational modules and plays an important role in biomedical research including radiomic, genetic and pathological studies. Due to the lack of criteria when reporting research findings of harmonisation studies, we strongly appeal that the researchers should follow and expand the checklist presented in this survey.

## Acknowledgement

This study was supported in part by the European Research Council Innovative Medicines Initiative (DRAGON[#], H2020-JTI-IMI2 101005122), the AI for Health Imaging Award (CHAIMELEON[##], H2020-SC1-FA-DTS-2019-1 952172), the UK Research and Innovation Future Leaders Fellowship (MR/V023799/1), the British Heart Foundation (Project Number: TG/18/5/34111, PG/16/78/32402), the SABER project supported by Boehringer Ingelheim Ltd, the European Union's Horizon 2020 research and innovation programme (ICOVID, 101016131), the Euskampus Foundation (COVID19 Resilience, Ref. COnfVID19), and the Basque Government (consolidated research group MATHMODE, Ref. IT1294-19, and 3KIA project from the ELKARTEK funding program, Ref. KK-2020/00049).

[#] DRAGON Consortium:

Xiaodan Xing[a], Ming Li[a], Scott Wagers[b], Rebecca Baker[c], Cosimo Nardi[d], Brice van Eeckhout[e], Paul Skipp[f], Pippa Powell[g], Miles Carroll[h], Alessandro Ruggiero[i], Muhunthan Thillai[i], Judith Babar[i], Evis Sala[i], William Murch[j], Julian Hiscox[k], Diana Baralle[l], Nicola Sverzellati[m]

[##] CHAIMELEON Consortium:

Ana Miguel Blanco[n], Fuensanta Bellvís Bataller[o], Mario Aznar[p], Amelia Suarez[p], Sergio Figueiras[q], Katharina Krischak[r], Monika Hierath[r], Yisroel Mirsky[s], Yuval Elovici[s], Jean Paul Beregi[t], Laure Fournier[t], Francesco Sardanelli[u], Tobias Penzkofer[v], Karine Seymour[w], Nacho Blanquer[x], Emanuele Neri[y], Andrea Laghi[z], Manuela França[aa], Ricard Martinez[ab]

[a] National Heart and Lung Institute, Imperial College London, London, UK
[b] BioSci Consulting, Maasmechelen, Belgium
[c] Clinical Data Interchange Standards Consortium, Austin, Texas, United States
[d] University of Florence, Firenze, Italy
[e] Medical Cloud Company, Liège, Belgium
[f] TopMD, Southampton, UK
[g] European Lung Foundation, Sheffield, UK
[h] Department of Health, Public Health England, London, UK
[i] Department of Radiology, University of Cambridge, Cambridge, UK
[j] Owlstone Medical, Cambridge, UK
[k] University of Liverpool, Liverpool, UK
[l] University of Southampton, Southampton, UK
[m] University of Parma, Parma, Italy
[n] Medical Imaging Department, Hospital Universitari i Politècnic La Fe, Valencia, Spain
[o] QUIBIM, Valencia, Spain
[p] Matical Innovation, Madrid, Spain
[q] Bahía Software, A Coruña, Spain
[r] European Institute for Biomedical Imaging Research, Vienna, Austria
[s] Ben Gurion University of the Negev, Be'er Sheva, Israel
[t] Le Collège des Enseignants en Radiologie de France, France
[u] Research Hospital Policlinico San Donato, Milan, Italy

[v] Charité – Universitätsmedizin Berlin, Berlin, Germany
[w] Medexprim, Labège, France
[x] Valencia Polytechnic University, Valencia, Spain
[y] University of Pisa, Pisa, Italy
[z] Sapienza University of Rome, Rome, Italy
[aa] The Centro Hospitalar Universitário do Porto, Portugal
[ab] University of Valencia, Valencia, Spain